%% file: meshgpt_arxiv.tex
\documentclass[10pt,twocolumn,letterpaper]{article}

\usepackage{graphicx}
\usepackage{amsmath}
\usepackage{amssymb,amsthm}
\usepackage{booktabs}
\usepackage{multirow}
\usepackage{makecell}
\usepackage[dvipsnames,table]{xcolor}
\usepackage{comment}
\usepackage{xspace}
\usepackage{tabularx}
\usepackage[table]{xcolor}
\usepackage{bbding}
\usepackage{pifont}

\usepackage[pagenumbers]{cvpr} 

\input{preamble}

\newcommand{\OURS}{MeshGPT}

\definecolor{cvprblue}{rgb}{0.21,0.49,0.74}
\usepackage[pagebackref,breaklinks,colorlinks,citecolor=cvprblue]{hyperref}

\title{\OURS: Generating Triangle Meshes with Decoder-Only Transformers}

\author{
Yawar Siddiqui$^{1}$~~~ 
Antonio Alliegro$^2$~~~
Alexey Artemov$^1$~~~\\
Tatiana Tommasi$^{2}$~~~
Daniele Sirigatti$^3$~~~
Vladislav Rosov$^3$~~~
Angela Dai$^1$~~~
Matthias Nie{\ss}ner$^1$~~~
\vspace{0.2cm} \\
Technical University of Munich$^1$~~~
Politecnico di Torino$^2$
AUDI AG$^3$
\vspace{0.2cm}
}

\begin{document}


\input{figures/teaser}
\maketitle
\input{sec/0_abstract}    
\input{sec/1_intro}

\input{sec/2_relatedwork}
\input{sec/3_method}
\input{sec/4_experiments}
\input{sec/5_conclusion}
\input{sec/6_acknowledge}
{\small
\bibliographystyle{ieeenat_fullname}
\bibliography{references}
}
\newpage

\begin{appendix}
\section*{Appendix}
\input{sec/7_appendix}
\end{appendix}


\end{document}

%% file: preamble.tex
\usepackage[dvipsnames]{xcolor}
\usepackage{bbm}
\usepackage{stfloats}

\newcommand{\mypara}[1]{\vspace{3pt}\noindent\textbf{#1}}

\newcommand{\argmin}{\operatornamewithlimits{argmin}}


%% file: figures/teaser.tex
\twocolumn[{%
	\renewcommand\twocolumn[1][]{#1}%
	\vspace{-30pt}
        \maketitle
	\begin{center}
        \vspace{-7mm}
		
        \includegraphics[width=\linewidth]{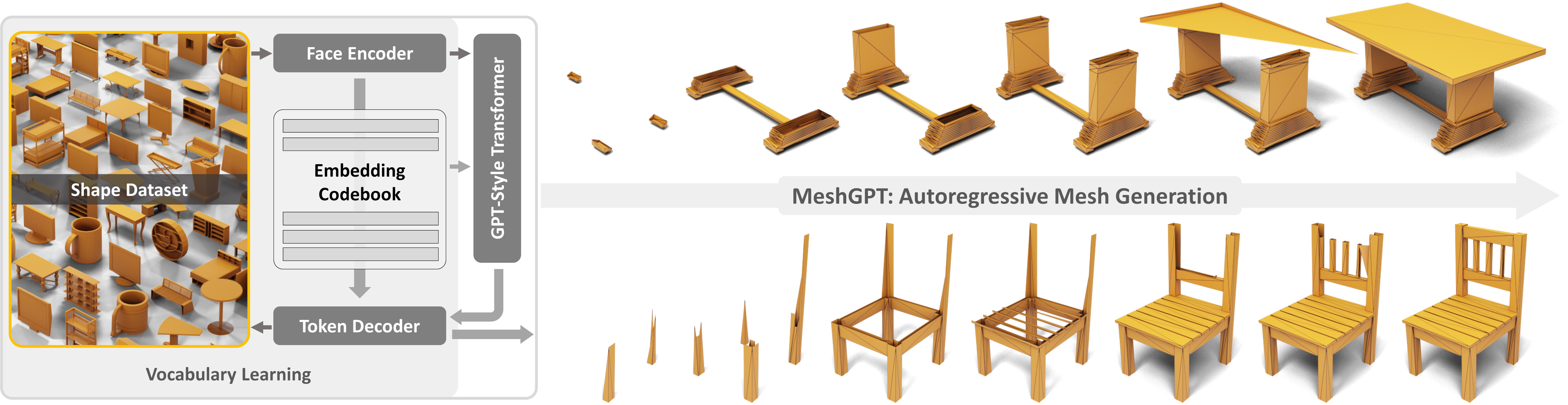}
            
		\captionof{figure}{ Our method creates triangle meshes by autoregressively sampling from a transformer model that has been trained to produce tokens from a learned geometric vocabulary. These tokens can then be decoded into the faces of a triangle mesh. Our method generates  clean, coherent, and compact meshes, characterized by sharp edges and high fidelity.
        }
		\label{fig:teaser}
	\end{center}    
}]

%% file: sec/0_abstract.tex
\begin{abstract}
We introduce \OURS{}, a new approach for generating triangle meshes that reflects the compactness typical of artist-created meshes, in contrast to dense triangle meshes extracted by iso-surfacing methods from neural fields. 
Inspired by recent advances in powerful large language models, we adopt a sequence-based approach to autoregressively generate triangle meshes as sequences of triangles.
We first learn a vocabulary of latent quantized embeddings, using graph convolutions, which inform these embeddings of the local mesh geometry and topology. 
These embeddings are sequenced and decoded into triangles by a decoder, ensuring that they can effectively reconstruct the mesh.
A transformer is then trained on this learned vocabulary to predict the index of the next embedding given previous embeddings. 
Once trained, our model can be autoregressively sampled to generate new triangle meshes, directly generating compact meshes with sharp edges, more closely imitating the efficient triangulation patterns of human-crafted meshes.
\OURS{} demonstrates a notable improvement over state of the art mesh generation methods, with a 9\% increase in shape coverage and a 30-point enhancement in FID scores across various categories.
\end{abstract}
\vspace{-4mm}

%% file: sec/1_intro.tex
\section{Introduction}
\label{sec:intro}

Triangle meshes are the main representation for 3D geometry in computer graphics. They are the predominant representation for 3D assets used in video games, movies, and virtual reality interfaces. Compared to alternative 3D shape representations such as point clouds or voxels, meshes provide a more coherent surface representation; they are more controllable, easier to manipulate, more compact, and fit directly into modern rendering pipelines, attaining high visual quality with far fewer primitives. In this paper, we tackle the task of automated generation of triangle meshes, streamlining the process of crafting 3D assets. 

Recently, 3D vision research has seen great interest in generative 3D models using representations such as voxels~\cite{wu2016learning, brock2016generative}, point clouds~\cite{luo2021diffusion, zeng2022lion, zhou20213d}, and neural fields~\cite{muller2023diffrf, li2023diffusion, liu2023meshdiffusion, gupta20233dgen, gao2022get3d}.
However, these representations must then be converted into meshes through a post-process for use in downstream applications, for instance by iso-surfacing with Marching Cubes~\cite{lorensen1998marching}.
Unfortunately, this results in dense, over-tessellated meshes that often exhibit oversmoothing and bumpy artifacts from the iso-surfacing, as shown in Figure~\ref{fig:topology}.
In contrast, artist-modeled 3D meshes are compact in representation, while maintaining sharp details with much fewer triangles.

Thus, we propose \OURS{}\footnote{\href{https://nihalsid.github.io/mesh-gpt/}{nihalsid.github.io/mesh-gpt}} to generate a mesh representation directly, as a set of triangles.
Inspired by powerful recent advances in generative models for language, we adopt a direct sequence generation approach to synthesize triangle meshes as sequences of triangles. 
Following text generation paradigms, we first learn a vocabulary of triangles.
Triangles are encoded into latent quantized embeddings through an encoder. 
To encourage learned triangle embeddings to maintain local geometric and topological features, we employ a graph convolutional encoder.
These triangle embeddings are then decoded by a ResNet~\cite{he2016deep} decoder that processes the sequence of tokens representing a triangle to produce its vertex coordinates.
We can then train a GPT-based architecture on this learned vocabulary to autoregressively produce sequences of triangles representing a mesh.
Experiments across multiple categories of the ShapeNet dataset demonstrate that our method significantly improves 3D mesh generation quality in comparison with state of the art, with an average 9\% increase in shape coverage and a 30-point improvement in FID scores.

\input{figures/compactness_topology}

\noindent In summary, our contributions are:
\begin{itemize}
    \item A new generative formulation for meshes as a sequence of triangles, tailoring a GPT-inspired decoder-only transformer, to produce compact meshes with sharp edges.
	\item Triangles are represented as a vocabulary of latent geometric tokens to enable coherent mesh generation in an autoregressive fashion.
\end{itemize}

%% file: figures/compactness_topology.tex
\begin{figure}[tp]
  \centering
   \includegraphics[width=0.98\linewidth]{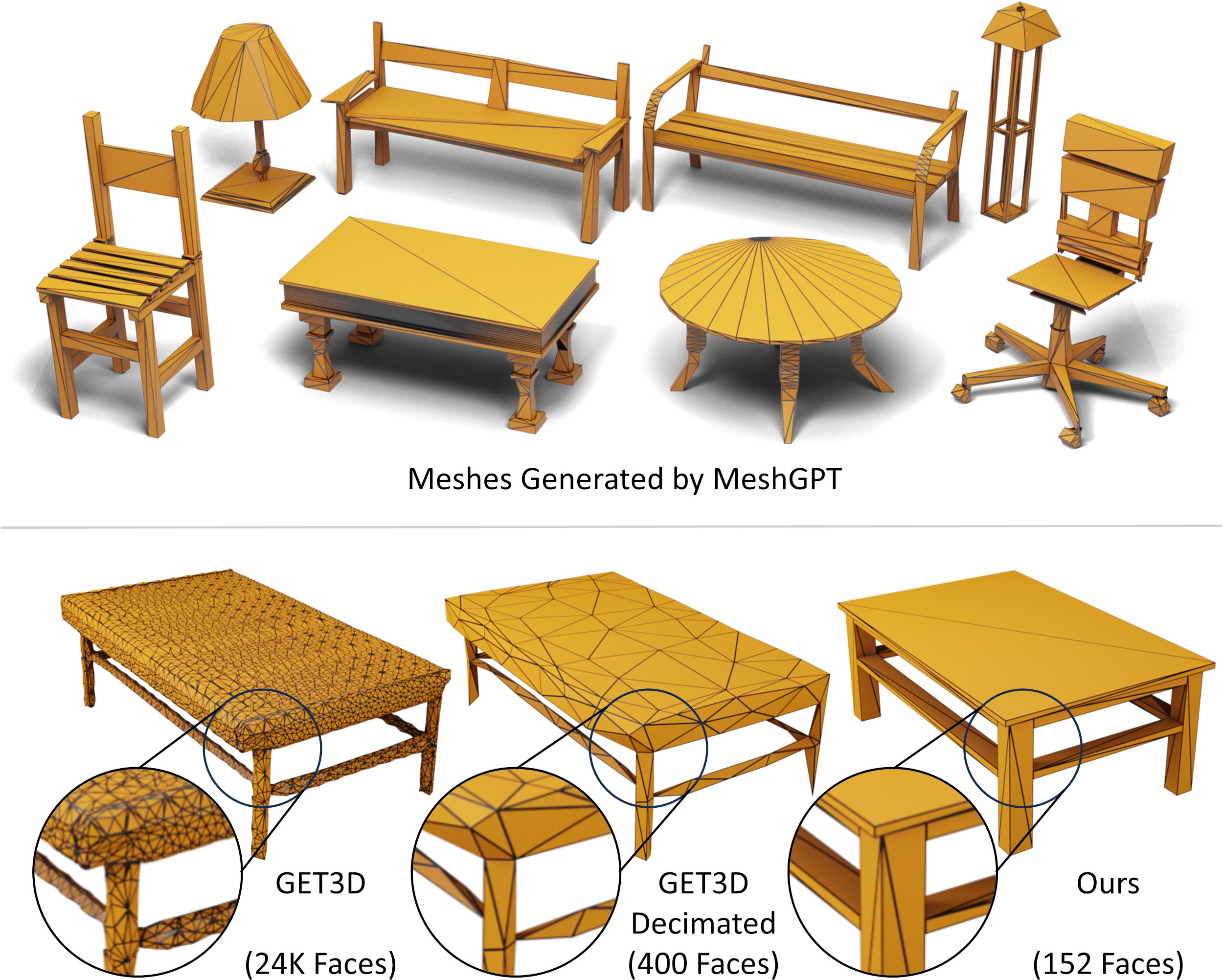}
   \caption{Meshes generated by our method (top) for chairs, tables, benches, and lamps when trained on ShapeNet~\cite{shapenet2015}. \OURS{} meshes tend to be compact, with the ability to represent both sharp details and curved boundaries. This contrasts with neural field-based approaches that yield dense triangulations not easily simplified through decimation (bottom).}
   \label{fig:topology}
   \vspace{-3mm}
\end{figure}

%% file: sec/2_relatedwork.tex
\section{Related Work}
\label{sec:related_work}
\mypara{Voxel-based 3D Shape Generation.}
Early shape generation approaches generated shapes as a grid of low-resolution voxels~\cite{wu2016learning,choy20163d,jimenez2016unsupervised, brock2016generative} or, more recently, as high-resolution grids using efficient representations such as Octrees~\cite{tatarchenko2017octree} and sparse voxels~\cite{schwarz2022voxgraf}, with generative models such as GANs~\cite{goodfellow2020generative}. These methods pioneered the extension of 2D generative techniques into the 3D domain. However, the voxel representation inherently constrains them with grid-like artifacts and high memory requirements, limiting their practical utility in capturing fine details and complex geometries.

\mypara{Point Cloud Generation.}
Methods in this category represent 3D shapes by point samples on their surfaces, aiming to learn point distributions across shape datasets. Early works involved GANs for synthesizing point locations~\cite{li2018point, valsesia2018learning, shu20193d} and latent shape codes~\cite{achlioptas2018learning}. Flow-based~\cite{yang2019pointflow} and gradient field-based models~\cite{cai2020learning} also yield impressive results. Recently, diffusion-based techniques have been adapted for point cloud generation~\cite{zhou20213d, zeng2022lion, nichol2022point}, showing competitive performance in shape generation. However, point clouds, while useful, are not the ideal format for downstream applications requiring 3D content, as converting them to meshes, which apart from being non-trivial~\cite{peng2021shape,sharp2020pointtrinet, nan2017polyfit,yang2018foldingnet}, can often fail to accurately reflect the characteristics of the underlying mesh datasets.

\mypara{Neural Implicit Fields.} Implicit representation of shapes as volumetric functions (\eg, signed distance functions) has become popular for encoding arbitrary topologies at any resolution~\cite{park2019deepsdf, mescheder2019occupancy}. Various implicit generative methods have shown impressive performance using adversarial~\cite{chen2019learning,shen2021deep} and diffusion-based~\cite{cheng2023sdfusion,gao2022get3d,muller2023diffrf} models. Diffusion-based neural field synthesis in MLP weight spaces~\cite{erkocc2023hyperdiffusion} and triplanes~\cite{shue20233d} have also been explored, alongside leveraging image-based models for optimizing NeRFs~\cite{poole2022dreamfusion,lin2023magic3d,jain2022zero,xu2023dream3d}. However, like point clouds, these methods require mesh conversion~\cite{doi1991efficient,lorensen1998marching,liao2018deep,remelli2020meshsdf,shen2021deep} for downstream applications, often leading to dense meshes that don’t capture the properties of the underlying datasets (\eg, edge lengths, dihedral angles). In contrast, we directly fit a generative model to triangulated meshes, explicitly modeling the training data, resulting in clean, compact and coherent meshes as outputs.

\mypara{3D Mesh Generation.} While several discriminative approaches capable of learning signals directly on mesh structure were proposed over the recent years~\cite{gong2019spiralnet++,sharp2022diffusionnet,milano2020primal,hu2022subdivision,lim2018simple,hanocka2019meshcnn,smirnov2021hodgenet}, direct mesh generation remains underexplored. Mesh generation has been approached with various learning-based methods~\cite{groueix2018papier,dai2019scan2mesh,chen2020bsp, nash2020polygen}. AtlasNet~\cite{groueix2018papier} and BSPNet~\cite{chen2020bsp}, for example, produce mesh patches and compact meshes through binary space partitioning, respectively. However, as we demonstrate in Sec.~\ref{sec:experiments}, these struggle with accurately capturing shape detail.

Closely related to our work, PolyGen~\cite{nash2020polygen} employs two autoregressively trained networks to create explicit mesh structures. In contrast, our method utilizes a single decoder-only network, representing triangles through learned tokens for a more streamlined generation process compared to PolyGen's separate vertex-and-face sequence approach. Additionally, we observe that PolyGen's vertex generator, oblivious to face generation, and the face generator, not exposed to the generated vertex distribution during training, exhibit limited robustness during inference.

%% file: sec/3_method.tex
\section{Method}
\label{sec:method}

Inspired by advancements in large language models, we develop a sequence-based approach to autoregressively generate triangle meshes as sequences of triangles. We first learn a vocabulary of geometric embeddings from a large collection of 3D object meshes, enabling triangles to be encoded to and decoded from this embedding. We then train a transformer for mesh generation as autoregressive next-index prediction over the learned vocabulary embeddings.

To learn the triangle vocabulary, we employ a graph convolution encoder operating on triangles of a mesh and their neighborhood to extract geometrically rich features that capture the intricate details of 3D shapes.  These features are quantized as embeddings of a codebook using residual quantization~\cite{martinez2014stacked, juang1982multiple}, effectively reducing sequence lengths of the mesh representation. These embeddings are sequenced and then decoded by a 1D ResNet~\cite{he2016deep} guided by a reconstruction loss. This phase lays the groundwork for the subsequent training of the transformer. 

We then train a GPT-style decoder-only transformer, which leverages these quantized geometric embeddings. Given a sequence of geometric embeddings extracted from the triangles of a mesh, the transformer is trained to predict the codebook index of the next embedding in the sequence. Once trained, the transformer can be auto-regressively sampled to predict sequences of embeddings. These embeddings can then be decoded to generate novel and diverse mesh structures that display efficient, irregular triangulations similar to human-crafted meshes.

\subsection{Learning Quantized Triangle Embeddings}
\label{sec:encoder_decoder}

\input{figures/reconstruction}

Autoregressive generative models, such as transformers, synthesize sequences of tokens where each new token is conditioned on previously generated tokens. For generating meshes using transformers, we must then define the ordering convention of generation, along with the tokens.

For sequence ordering, Polygen~\cite{nash2020polygen} suggests a convention where faces are ordered based on their lowest vertex index, followed by the next lowest, and so forth. 
Vertices are sorted in $z$-$y$-$x$ order ($z$ representing the vertical axis), progressing from lowest to highest.
Within each face, indices are cyclically permuted to place the lowest index first.
In our method, we also adopt this sequencing approach.

To define the tokens to generate, we consider a practical approach to represent a mesh $\mathcal{M}$ for autoregressive generation: a sequence of triangles,
\begin{equation}
    \mathcal{M} := (f_1, f_2, f_3, \ldots, f_N), 
\end{equation}
with $N$ faces (triangles), $f_i \in \mathbb{R}^{n_\text{in}}$ having $n_\text{in}$ features. 
A simple approach to describe each triangle is as its three vertices, comprising nine total coordinates.  Upon discretization, these coordinates can be treated as tokens. The sequence length in this case would be $9N$.

However, we observe two major challenges when using coordinates directly as tokens. First, the sequence lengths become excessively long, as each face is represented by nine values. This length does not scale well with transformer architectures, which often have limited context windows. Second, representing discrete positions of a triangle as tokens fails to capture geometric patterns effectively. This is because such a representation lacks information about neighboring triangles and does not incorporate any priors from mesh distributions.

To address the aforementioned challenges, we propose to learn geometric embeddings from a collection of triangular meshes, utilizing an encoder-decoder architecture with residual vector quantization at its bottleneck (Fig.~\ref{fig:method_vq}). 

The network's encoder $E$ employs graph convolutions on mesh faces, where each face forms a node and neighboring faces are connected by undirected edges. The input face node features are comprised of the nine positionally encoded coordinates of its vertices, face normal, angles between its edges, and area. These features undergo processing through a stack of SAGEConv~\cite{hamilton2017inductive} layers, extracting a feature vector for each face. This graph convolutional approach enables the extraction of geometrically enriched features $z_i \in \mathbb{R}^{n_\text{z}} $ for each face,
\begin{equation}
    \mathbf{Z} = (z_1, z_2, \ldots, z_N) = E(\mathcal{M}), 
\end{equation}
fusing neighborhood information into the learned embeddings.

For quantization, we employ residual vector quantization (RQ)~\cite{martinez2014stacked}. We found that using a single code per face is insufficient for accurate reconstruction. Instead, we use a stack of D codes per face. Further, we find that instead of directly using D codes per face, it is more effective to first divide the feature channels among the vertices, aggregate the features by shared vertex indices, and then quantize these vertex-based features, giving $\frac{\text{D}}{3}$ codes per vertex, and therefore effectively D codes per face. This leads to sequences that are easier to learn for the transformer trained subsequently (see Tab.~\ref{tab:ablations} and Fig.~\ref{fig:results_ablation} for comparison).
Formally, given a codebook $\mathcal{C}$, RQ with depth D represents features $\mathbf{Z}$ as 
\begin{equation}
\label{eq:token_sequence}
    \mathbf{T} = (t_1, t_2, \ldots, t_N) = \text{RQ}\,(\mathbf{Z}; \mathcal{C}, D),
\end{equation} 
\begin{equation}
    t_i = (t_{i}^{1}, t_{i}^{2}, \ldots, t_{i}^{D}),
\end{equation} 
where $t_i$ is a stack of tokens, each token $t_{i}^{d}$ being an index to an embedding $\mathbf{e}(t_{i}^{d})$ in the codebook $\mathcal{C}$. 

The decoder then decodes the quantized face embeddings to triangles. First, the stack of D features is reduced to a single feature per face through summation across embeddings and concatenation across vertices, 
\begin{equation}
    \mathbf{\hat{Z}} = (\hat{z}_1, \ldots, \hat{z}_N), \;\text{with}\;
    \hat{z}_i = \oplus_{v=0}^{2}\sum_{d=1}^{\frac{D}{3}}{ \,\mathbf{e}(t_{i}^{3.v + d})}.
\end{equation} 
The face embeddings are arranged in the previously described order, and a 1D ResNet34 decoding head $G$ processes the resulting sequence to output the reconstructed mesh $\hat{\mathcal{M}} = G(\mathbf{\hat{Z}})$ with $9$ coordinates representing each face. We observe that predicting these coordinates as discrete variables, i.e. as a probability distribution over a set of discrete values, leads to a more accurate reconstruction compared to regressing them as real values (Fig.~\ref{fig:discrete_vs_continuous}). A cross-entropy loss on the discrete mesh coordinates and a commitment loss for the embeddings guides the reconstruction process. More details can be found in supplementary.

\input{figures/discrete_vs_continuous}

After training, the graph encoder $E$ and codebook $\mathcal{C}$ are incorporated into the transformer training, using $\mathbf{T}$ from Eq.~\ref{eq:token_sequence} as the token sequence. With $|\mathbf{T}| = DN$, this sequence is more concise than the naive $9N$-length tokenization when $D < 9$. Thus, we obtain geometrically rich embeddings with shorter sequence lengths, overcoming our initial challenges and paving the way for efficient mesh generation.

\subsection{Mesh Generation with Transformers}

\input{figures/sequencing}

We employ a decoder-only transformer architecture from the GPT family of models to predict meshes as sequences of indices from the learned codebook in Sec.~\ref{sec:encoder_decoder}. The input to this transformer consists of embeddings $\mathbf{e}(t_{i}^{d})$ extracted from the mesh $\mathcal{M}$ using the GraphConv encoder $E$ and quantized using RQ (Eq.~\ref{eq:token_sequence}). The embeddings are prefixed and suffixed with a learned start and end embedding. Additionally, learned discrete positional encodings are added, indicating the position of each face in the sequence and the index of each embedding within the face. The features then pass through a stack of multiheaded self-attention layers, where the transformer is trained to predict the codebook index of the next embedding in the sequence (Fig.~\ref{fig:method_transformer}). Essentially, we maximize the log probability of the training sequences with respect to the transformer parameters $\theta$,   
\begin{equation}
    \prod_{i=1}^{N}\prod_{d=1}^{D} p\,(t_{i}^{d}\,|\,\mathbf{e}(t_{<i}^{d}),\,\mathbf{e}(t_{i}^{<d});\,\theta).
\end{equation}

Once the transformer is trained, it can autoregressively generate a sequence of tokens, starting with a start token and continuing until a stop token is encountered using beam sampling. The codebook embeddings indexed by this sequence of tokens is then decoded by decoder $G$ to produce the generated mesh. As this output initially forms a `triangle soup' with duplicate vertices for neighboring faces, we apply a simple post-processing operation to merge close vertices (e.g., with  MeshLab), to yield the final mesh.

\subsection{Implementation Details}
In learning the triangle vocabulary, our residual quantization layer features a depth of 2, yielding
$D=6$ embeddings per face, each with dimension 192. The codebook is dynamically updated using an exponential moving average of the clustered features. Following~\cite{lee2022autoregressive}, we incorporate stochastic sampling of codes and employ a shared codebook across all levels. The decoder predicts the coordinates of the faces across 128 classes, resulting in a discretization of space to $128^3$ possible values. This encoder-decoder network is trained using 2 A100 GPUs for $\approx$ 2 days.

For our transformer, we use a GPT2-medium model, equipped with a context window of up to 4608 embeddings. The model is trained on 4 A100 GPUs, for $\approx$ 5 days.

Both the encoder-decoder network and the transformer are written using the Pytorch ~\cite{NEURIPS2019_9015} and are trained utilizing the ADAM optimizer~\cite{kingma2014adam}. We set the learning rate at $1\times10^{-4}$ and use an effective batch size of 64.

%% file: figures/reconstruction.tex
\begin{figure}[htp]
  \centering
   \includegraphics[width=\linewidth]{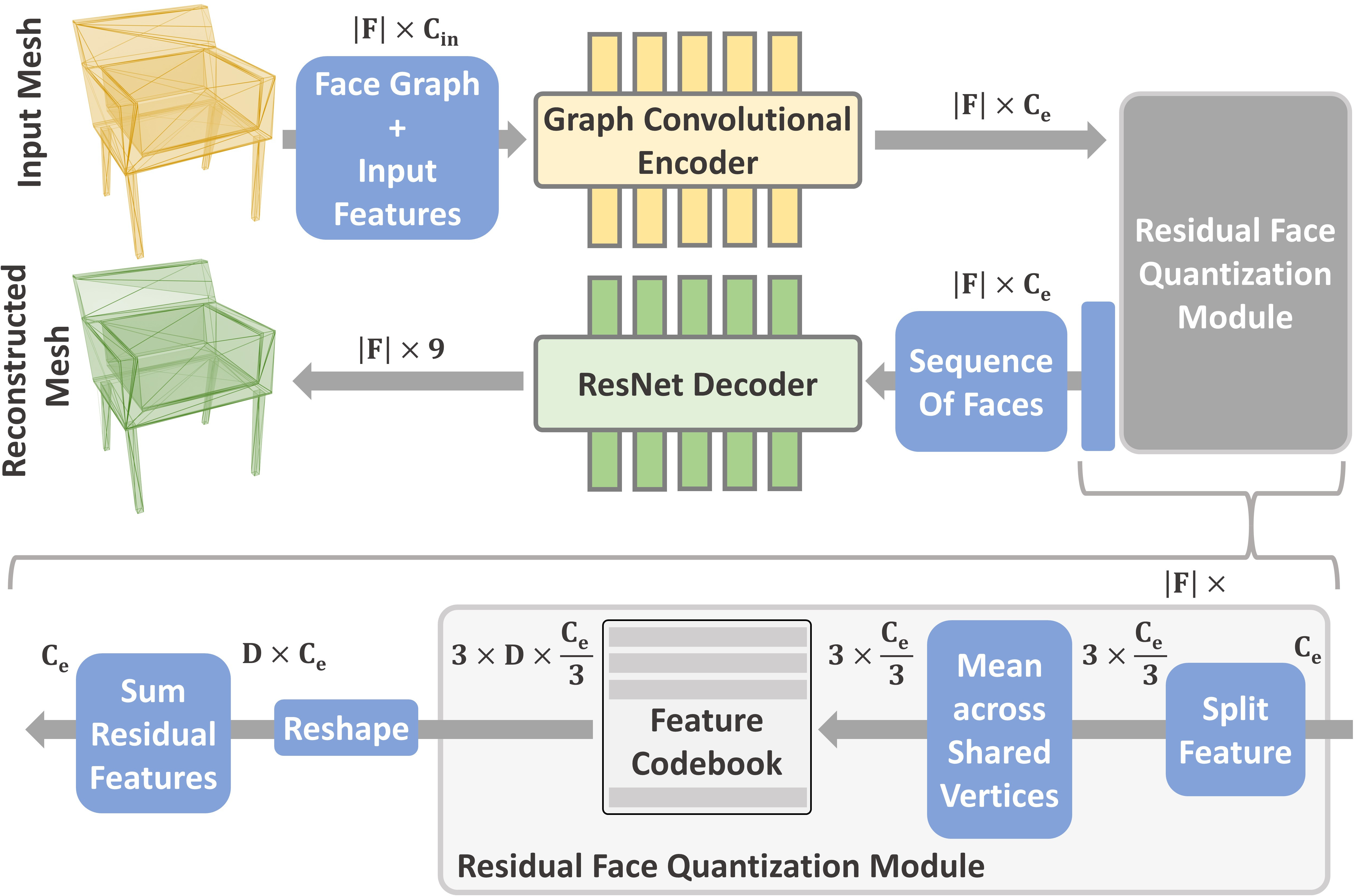}
   \caption{
   We employ a graph convolutional encoder to process mesh faces, leveraging geometric neighborhood information to capture strong features representing intricate details of 3D shapes. 
   These features are then quantized into codebook embeddings using residual quantization~\cite{juang1982multiple, martinez2014stacked}. In contrast to naive vector quantization,  this ensures better reconstruction quality. The quantized embeddings are subsequently sequenced and decoded through a 1D ResNet~\cite{he2016deep}, guided by a reconstruction loss.
   }
   \label{fig:method_vq}
\end{figure}

%% file: figures/discrete_vs_continuous.tex
\begin{figure}[htp]
  \centering
   \includegraphics[width=0.90\linewidth]{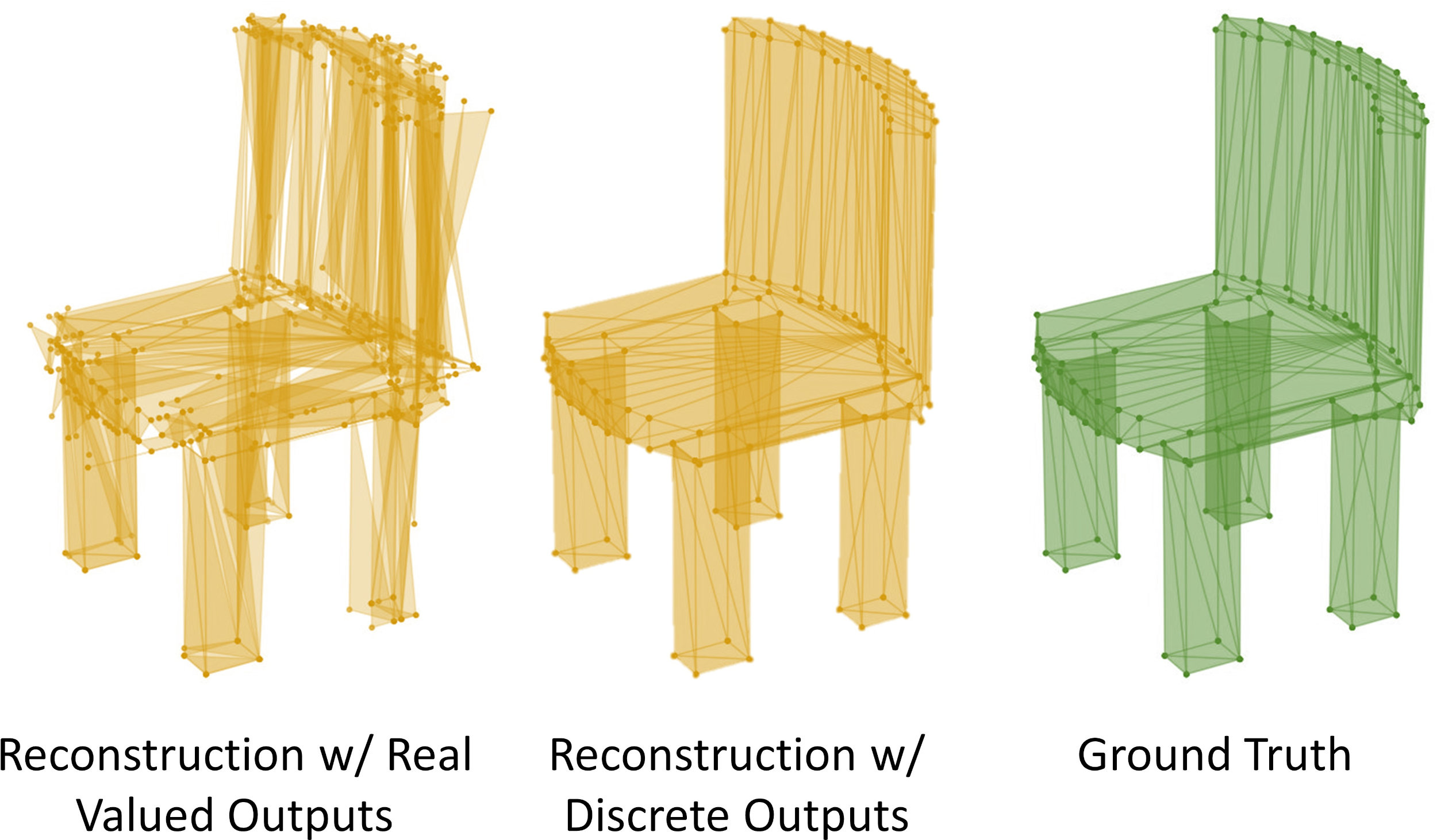}
   \caption{Our method utilizes a ResNet~\cite{he2016deep} decoder that outputs mesh faces as a distribution over discretized coordinate values (center), as opposed to regression of continuous values (left). This significantly reduces floating face artifacts, leading to reconstructions that more closely resemble the ground truth (right).}
   \label{fig:discrete_vs_continuous}
\end{figure}

%% file: figures/sequencing.tex
\begin{figure}[htbp]
  \centering
   \includegraphics[width=\linewidth]{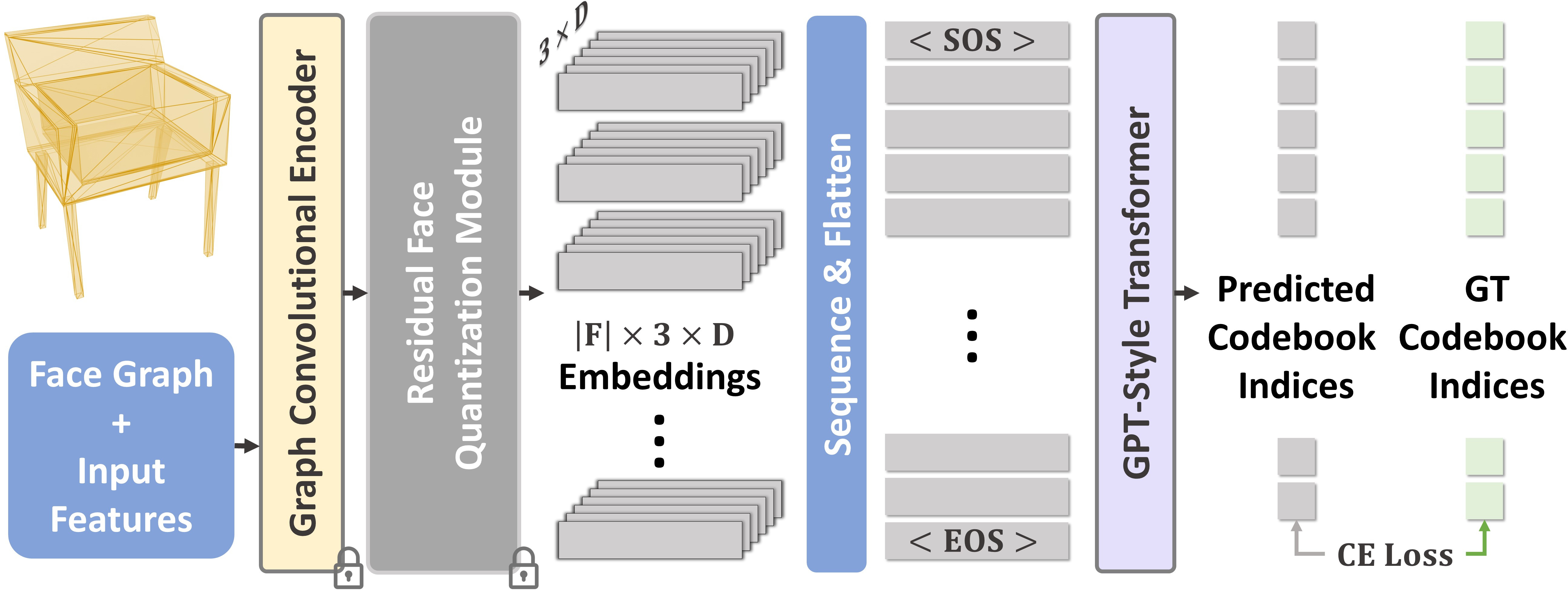}
   \caption{We employ a transformer to generate mesh sequences as token indices from a pre-learned codebook vocabulary. During training, a graph encoder extracts features from mesh faces, which are quantized into a set of face embeddings. These embeddings are flattened, bookended with start and end tokens, and fed into a GPT-style transformer. This decoder predicts the subsequent codebook index for each embedding, optimized via cross-entropy loss.}
   \label{fig:method_transformer}
\end{figure}

%% file: sec/4_experiments.tex
\section{Experiments}
\label{sec:experiments}
\subsection{Dataset and Metrics}
\mypara{Data.}
We present our results on the ShapeNetV2 dataset. Both the encoder-decoder network and the GPT model are trained across all 55 categories of this dataset. Additionally, we fine-tune the GPT model specifically on four categories: Chair, Table, Bench, and Lamp. The results are reported on these categories. During training, we employ augmentation techniques including random shifts and random scaling to enhance the diversity of the training meshes. Similar to Polygen~\cite{nash2020polygen}, we also apply planar decimation to further augment the shapes. To ensure that the entire mesh fits into the transformer's context window, we select only those meshes for training that have fewer than 800 faces post-decimation. Detailed information regarding the augmentation processes, decimation techniques, and data splits are provided in the supplementary material.

\mypara{Metrics.} Evaluating the unconditional synthesis of 3D shapes presents challenges due to the absence of direct ground truth correspondence. Hence, we utilize established metrics for assessment, consistent with previous works~\cite{zeng2022lion,luo2021diffusion,zhou20213d}. These include Minimum Matching Distance (MMD), Coverage (COV), and 1-Nearest-Neighbor Accuracy (1-NNA). For MMD, lower is better; for COV, higher is better; for 1-NNA, 50\% is the optimal. We use a Chamfer Distance (CD) distance measure for computing these metrics in 3D. More details about these metrics can be found in the supplementary.

The aforementioned metrics effectively measure the quality of shapes but do not address the visual similarity of the generated meshes to the real distribution. To assess this aspect, we render both the generated meshes and the ShapeNet meshes as images from eight different viewpoints using Blender, applying a metallic material to emphasize the geometric structures. Subsequently, we calculate the FID (Fréchet Inception Distance) and KID (Kernel Inception Distance) scores for these image sets. For both FID and KID, lower scores indicate better performance. We further report compactness as the average number of vertices and faces in the generated meshes. 

\subsection{Results}

\input{tables/results_main}

\input{figures/qualitative_chair_table}

\input{figures/qualitative_bench_lamp}

We benchmark our approach against leading mesh generation methods: Polygen~\cite{nash2020polygen}, which generates polygonal meshes by first generating vertices followed by faces conditioned on the vertices; BSPNet~\cite{chen2020bsp}, which represents a mesh through convex decompositions; and AtlasNet~\cite{groueix2018papier}, which represents a 3D mesh as a deformation of multiple 2D planes. We additionally compare with a state-of-the-art neural field-based method, GET3D~\cite{gao2022get3d}, that creates shapes as 3D signed distance fields (SDFs) from which a mesh is extracted by differentiable marching tetrahedra. For BSPNet and AtlasNet, which are built on autoencoder backbones, we follow \cite{achlioptas2018learning} to fit a Gaussian mixture model with 32 components to enable unconditional sampling of shapes.

As shown in Fig.~\ref{fig:results_chair_table}, Fig.~\ref{fig:results_bench_lamp} and Tab.~\ref{tab:quantative_evaluation}, our method outperforms all baselines in all four categories. Our method can generate sharp and compact meshes with high geometric details. Compared to Polygen, our approach creates shapes with more intricate details. Additionally, Polygen's separate training for vertex and face models, with the latter only exposed to ground truth vertex distributions, makes it more susceptible to error accumulation during inference. AtlasNet often suffers from folding artifacts, resulting in lower diversity and shape quality. BSPNet’s use of BSP tree of planes tends to produce blocky shapes with unusual triangulation patterns. GET3D generates good high-level shape structures, but over-triangulated and with imperfect flat surfaces. Simplifying GET3D-generated meshes with algorithms such as QEM~\cite{garland1997surface} results in a loss of fine structures.

\mypara{User Study.} We further conducted a user study, in Tab.~\ref{tab:user_study}, to assess generated mesh quality.
49 participants were shown pairs of four meshes, randomly selected from our method and each baseline method. Additionally, users were presented with ground truth ShapeNet meshes for comparison. Participants were asked their preference between our method and the baseline in terms of both overall shape quality and similarity of triangulation patterns to the ground truth meshes. 
This resulted in 784 total question responses. 
Our method was significantly preferred over AtlasNet, Polygen, and BSPNet in both shape and triangulation quality. Moreover, a majority of users (68\%) favored our method over neural field-based GET3D in shape quality, with even higher preference (73\%) for triangulation quality. This underscores our ability to generate high-quality meshes that align with human users’ preferences. 
Further user study details are provided in the supplemental.

\input{figures/shape_novelty}

\input{figures/shape_completion}

\input{tables/results_user_study}
\input{tables/results_ablation}

\mypara{Shape Novelty Analysis.} We investigate whether our method can generate novel shapes that extend beyond the training dataset, ensuring the model is not merely retrieving existing shapes. Following the methodology in previous studies \cite{erkocc2023hyperdiffusion, hui2022neural}, we generate 500 shapes using our model. For each generated shape, we identify the top three nearest neighbors from the training set based on Chamfer Distance (CD). To ensure a fair distance computation, accounting for potential discrepancies due to scale or shift in the augmented generations, we normalize all generated and train shapes to be centered within $[0,1]^3$.

Fig.~\ref{fig:shape_novelty} displays the most similar shapes from the train set corresponding to a sample generated by our model. We conduct a detailed analysis of the shape similarity distribution between the retrieved and generated shapes on the Chair category in Fig.~\ref{fig:shape_novelty}. The CD distribution reveals that our method not only covers shapes in the training set, indicated by low CD values, but also successfully generates novel and realistic-looking shapes, indicated by high CD values.
In the supplemental, we present further analysis of the novelty of all meshes generated by our method, which are featured in the figures of this paper. 

\mypara{Shape Completion.} Our model can infer multiple possible completions for a given partial shape, leveraging its probabilistic nature to generate diverse shape hypotheses. Fig.~\ref{fig:shape_completion} illustrates examples of chair and table completions.

\subsubsection{Ablations}
In Tab.~\ref{tab:ablations}, we show a set of ablations on the task of unconditional mesh generation on ShapeNet Chair category.  
Further ablations are detailed in the supplementary.

\input{figures/ablation}

\mypara{Do learned geometric embeddings help?} Using our geometric embeddings in vocabulary learning significantly improves over naive coordinate tokenization (w/o Learned Tokens), as shown in Tab.~\ref{tab:ablations} and Fig.~\ref{tab:ablations}.

\mypara{Does sequence length compression help?} As evidenced by Tab.~\ref{tab:ablations} and Fig.~\ref{tab:ablations}, a model with a shorter sequence length performs better than without (w/o Sequence Compression), as shorter sequence lengths fit transformer context windows better. Visually, with longer sequences, shapes exhibit repeating structures due to limited context. 

\mypara{What is the effect of aggregation and quantization across vertex indices instead of faces?}
As discussed in Section~\ref{sec:method}, an alternative to having embeddings aggregated and quantized across vertex indices, is to simply have the same number of embeddings directly per face (w/o per Vertex Quantization). In Tab.~\ref{tab:ablations}, we observe that this makes sequences much harder to learn with the transformer. 

\mypara{Do features from graph convolutional encoder help in mesh generation?} An alternative to using embeddings from the graph encoder and the codebook is to only use the codebook indices of these tokens as input to the transformer, and let the transformer learn the discrete token embeddings (w/o Encoder Features). While the transformer is able to still learn meaningful embeddings, these are still not as effective as using graph encoder features. 

\mypara{What is the effect of large-scale shape pretraining?} Tab.~\ref{tab:ablations} shows that training only on shapes from individual categories (w/o Pretraining) leads to overfitting and subobtimal performance, in contrast to pre-training our GPT transformer on all ShapeNet train shapes. 

\mypara{Limitations.}
\OURS{} significantly advances direct mesh generation but faces several limitations.  Its autoregressive nature leads to slower sampling performance, with mesh generation times taking 30 to 90 seconds. Despite our learned tokenization approach reducing sequence lengths, which suffices for single object generation, it may not be as effective for scene-scale generation, suggesting an area for future enhancement.  Moreover, our current computational resources limit us to using a GPT2-medium transformer, which is smaller than more sophisticated models like Llama2~\cite{touvron2023llama}. Given that larger language models benefit from increased data and computational power, expanding these resources could significantly boost MeshGPT’s performance and capabilities.

%% file: tables/results_main.tex
{
\begin{table}[tp]
  \begin{center}
    \small
    \resizebox{\linewidth}{!}{
      \begin{tabular}{p{7mm}lrrrrrrr}
        \toprule
        Class               & \multicolumn{1}{c}{Method}                          & COV$\uparrow$  & MMD$\downarrow$ & 1-NNA          & FID$\downarrow$ & KID$\downarrow$  & $\mid$V$\mid$   & $\mid$F$\mid$   \\
        \midrule

        \multirow{6}{*}{Chair} & AtlasNet~\cite{groueix2018papier}                   & 9.03           & 4.05            & 95.13          & 170.71          & 0.169          & 2500  & 4050  \\

                               & BSPNet~\cite{chen2020bsp}                           & 16.48          & 3.62            & 91.75          & 46.73           & 0.030          & 673   & 1165  \\

                               & Polygen~\cite{nash2020polygen}                      & 31.22          & 4.41            & 93.56          & 61.10           & 0.043          & 248   & 603   \\
                               & GET3D~\cite{gao2022get3d}                           & 40.85          & 3.56            & 83.04          & 81.45           & 0.054          & 13725 & 27457 \\
                               & GET3D* & 38.75          & 3.57            & 84.07          & 78.29           & 0.065          & 199   & 399   \\
        \cmidrule{2-9}
                               & \OURS{}                                             & \textbf{43.28} & \textbf{3.29}   & \textbf{75.51} & \textbf{18.46}  & \textbf{0.010} & 125   & 228   \\
        \bottomrule

        \multirow{6}{*}{Table} & AtlasNet~\cite{groueix2018papier}                   & 7.16           & 3.85            & 96.30          & 161.38          & 0.150          & 2500  & 4050  \\

                               & BSPNet~\cite{chen2020bsp}                           & 16.83          & 3.14            & 93.58          & 30.78           & 0.017          & 420   & 699   \\

                               & Polygen~\cite{nash2020polygen}                      & 32.99          & 3.00            & 88.65          & 38.53           & 0.029          & 147   & 454   \\
                               & GET3D~\cite{gao2022get3d}                           & 41.70          & 2.78            & 85.54          & 93.93           & 0.076          & 13767 & 27537 \\
                               & GET3D* & 37.95          & 2.85            & 81.93          & 50.46           & 0.037          & 199   & 399   \\
        \cmidrule{2-9}
                               & \OURS{}                                             & \textbf{45.68} & \textbf{2.36}   & \textbf{72.88} & \textbf{6.24}   & \textbf{0.002} & 99    & 187   \\
        \bottomrule

        \multirow{6}{*}{Bench} & AtlasNet~\cite{groueix2018papier}                   & 20.53          & 2.47            & 90.58          & 189.39          & 0.163          & 2500  & 4050  \\

                               & BSPNet~\cite{chen2020bsp}                           & 28.74          & 2.05            & 88.44          & 59.11           & 0.030          & 457   & 756   \\

                               & Polygen~\cite{nash2020polygen}                      & 51.92          & 1.97            & 76.98          & 49.34           & 0.031          & 172   & 430   \\
        \cmidrule{2-9}
                               & \OURS{}                                             & \textbf{55.23} & \textbf{1.44}   & \textbf{68.24} & \textbf{8.72}   & \textbf{0.001} & 159   & 291   \\
        \bottomrule

        \multirow{6}{*}{Lamp}  & AtlasNet~\cite{groueix2018papier}                   & 19.97          & 4.68            & 91.85          & 177.91          & 0.139          & 2500  & 4050  \\

                               & BSPNet~\cite{chen2020bsp}                           & 18.38          & 5.32            & 93.13          & 112.65          & 0.077          & 587   & 1011  \\

                               & Polygen~\cite{nash2020polygen}                      & 47.86          & 4.18            & 81.42          & 52.48           & 0.025          & 185   & 558   \\
        \cmidrule{2-9}
                               & \OURS{}                                             & \textbf{53.88} & \textbf{3.94}   & \textbf{65.73} & \textbf{19.91}  & \textbf{0.004} & 150   & 288   \\
        \bottomrule
      \end{tabular}}
    \caption{Quantitative comparison on the task of unconditional mesh generation on a subset of categories from the ShapeNet~\cite{shapenet2015} dataset. GET3D* refers to meshes simplified to 400 faces using QEM~\cite{garland1997surface}. MMD values are multiplied by $10^3$. We outperform the baselines on shape quality, visual and compactness metrics.}
    \label{tab:quantative_evaluation}
    \vspace{-8mm}
  \end{center}

\end{table}
}

%% file: figures/qualitative_chair_table.tex
\begin{figure*}[htp]
  \centering
   \includegraphics[width=\linewidth]{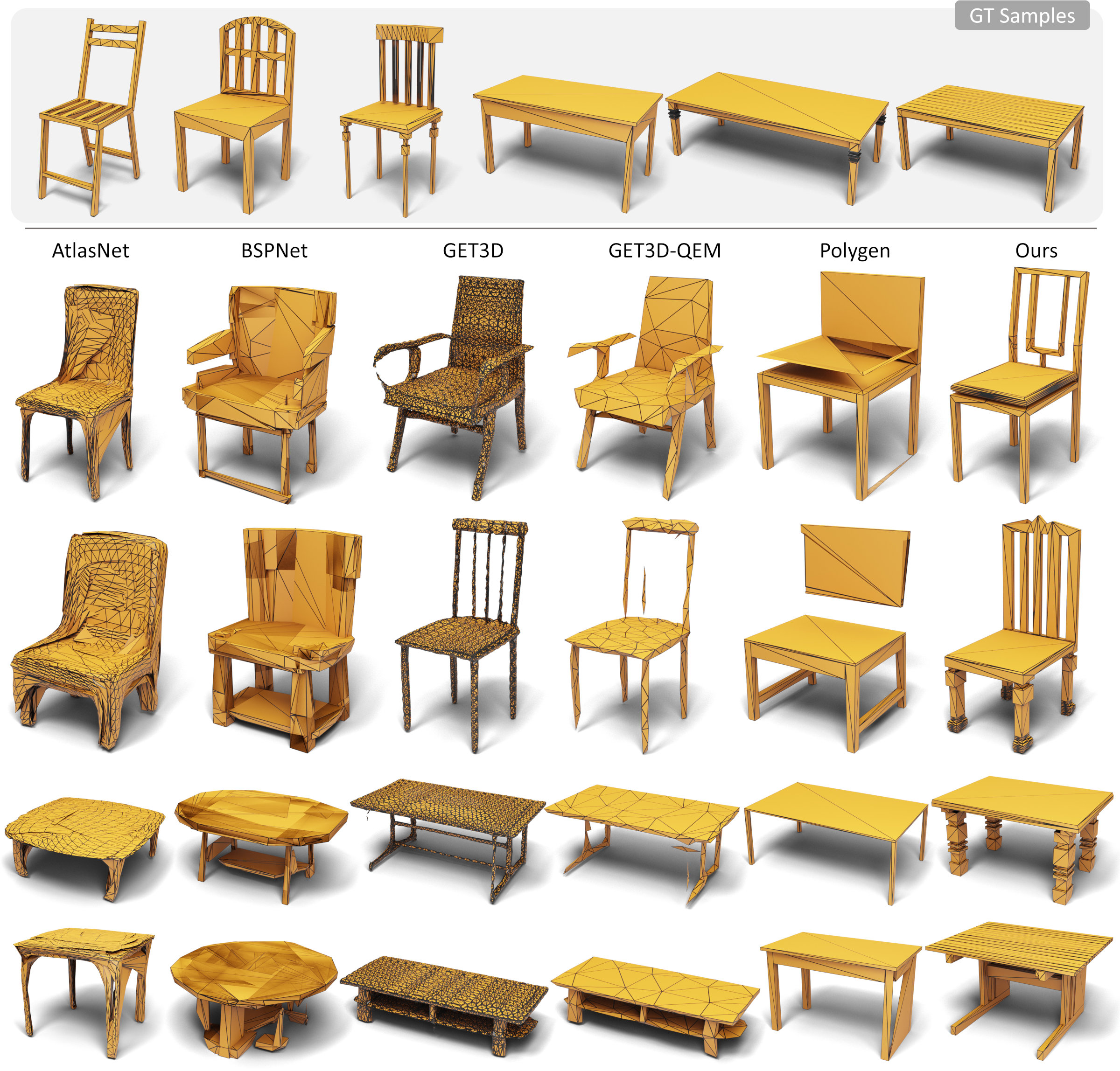}
   \caption{Qualitative comparison of Chair and Table meshes from ShapeNet~\cite{shapenet2015}. Our approach produces compact meshes with sharp geometric details. In contrast, baselines often either miss these details, produce over-triangulated meshes, or output too simplistic shapes.}
   \label{fig:results_chair_table}
   \vspace{-5mm}
\end{figure*}

%% file: figures/qualitative_bench_lamp.tex
\begin{figure}[htp]
  \centering
   \includegraphics[width=\linewidth]{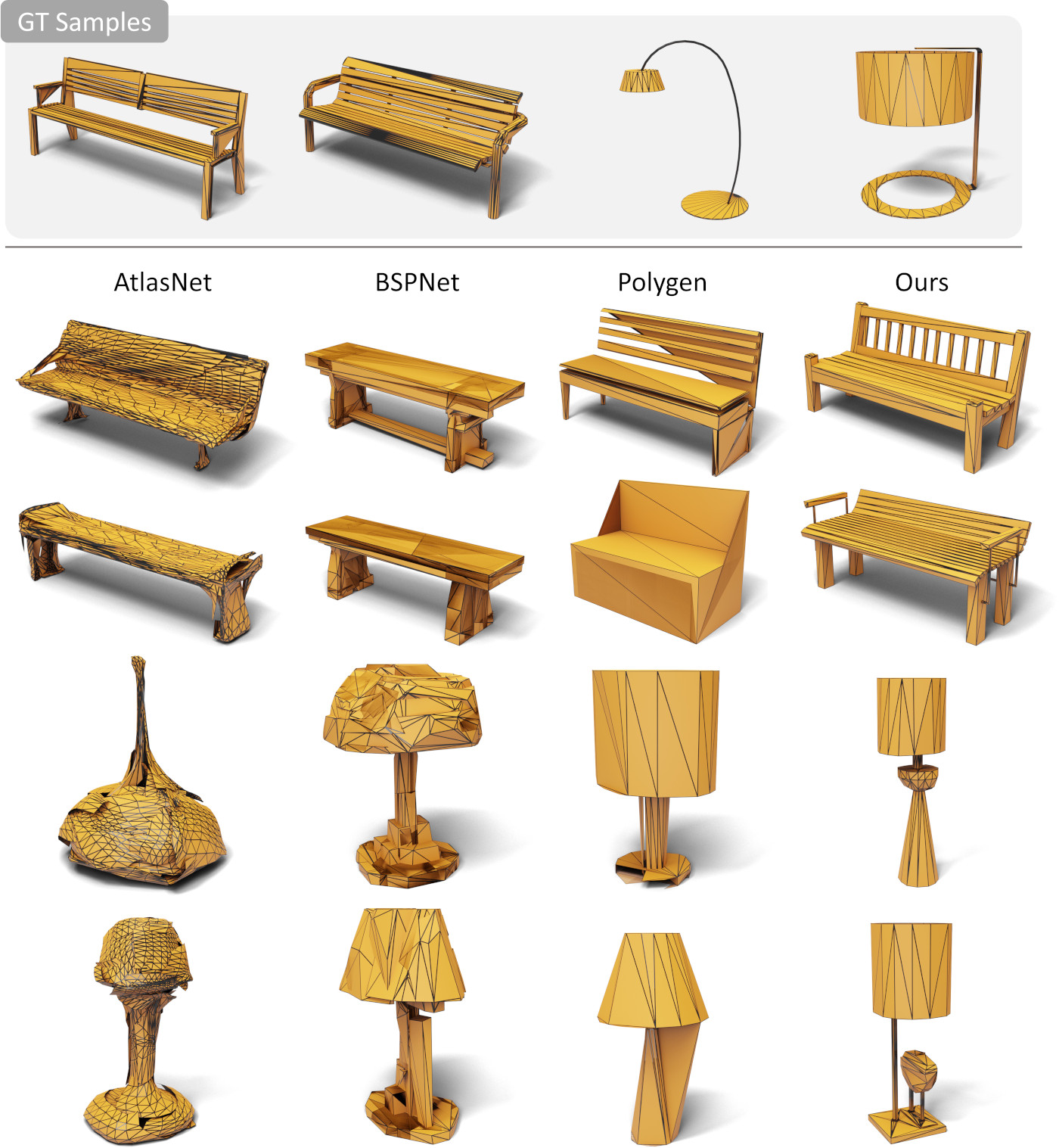}
   \caption{Qualitative comparison of Bench and Lamp meshes from the ShapeNet~\cite{shapenet2015} dataset. Compared to baselines, our method produces valid meshes with high geometric fidelity.}
   \label{fig:results_bench_lamp}
   \vspace{-3mm}
\end{figure}

%% file: figures/shape_novelty.tex
\begin{figure}[htp]
  \centering
   \includegraphics[width=\linewidth]{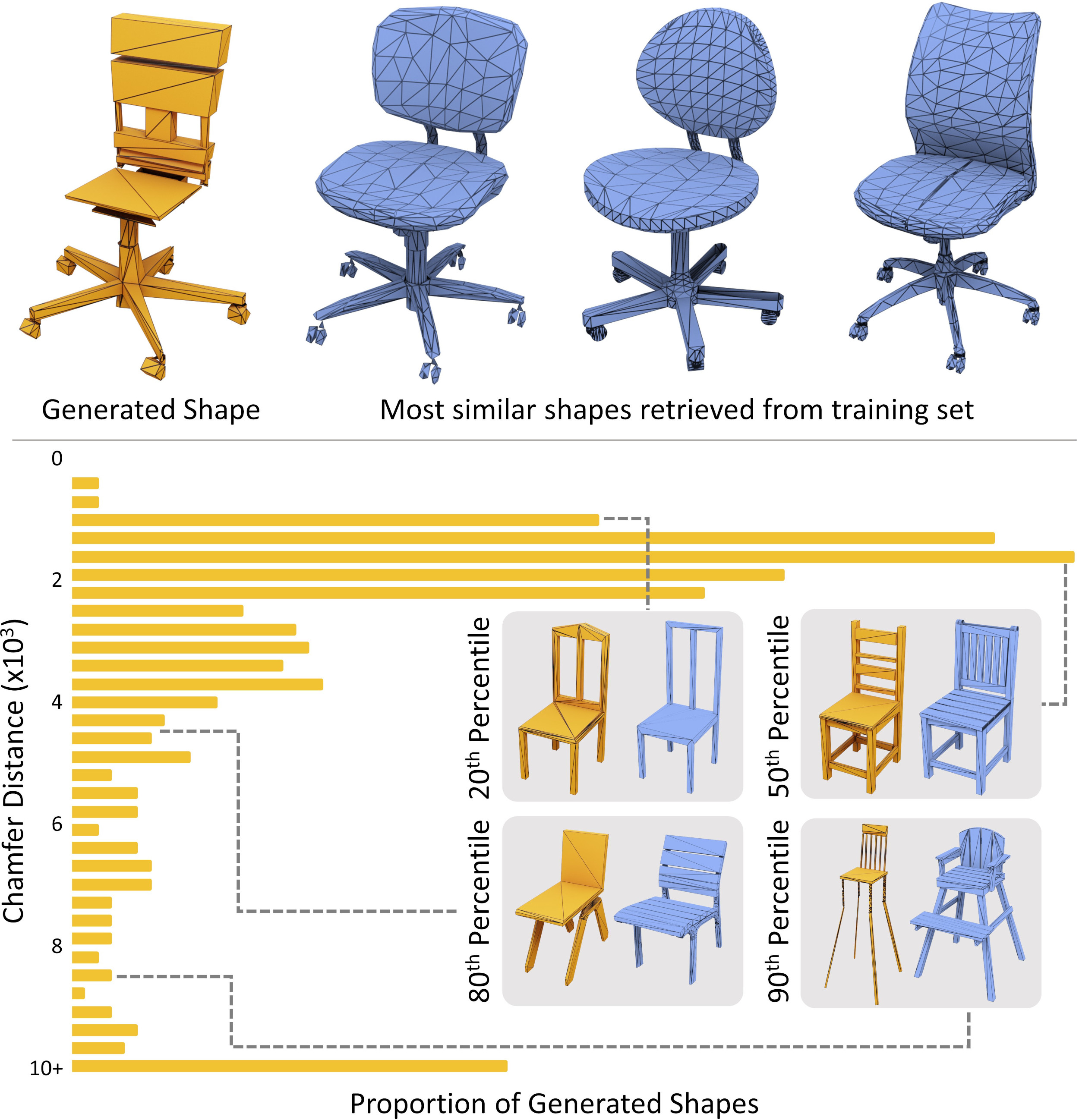}
   \caption{Shape novelty analysis on ShapeNet~\cite{shapenet2015} chair category. We show the 3 nearest neighbors in terms of Chamfer Distance (CD) for a generated shape (top). We also plot the distribution of 500 generated chair samples from our method and their closeness to training distribution. Our method can generate shapes that are similar (low CD) as well as different (high CD) from the training distribution, with shapes at the $50^\text{th}$ percentile looking different from closest train shape.}
   \label{fig:shape_novelty}
   \vspace{-3mm}
\end{figure}

%% file: figures/shape_completion.tex
\begin{figure}[htp]
  \centering
   \includegraphics[width=\linewidth]{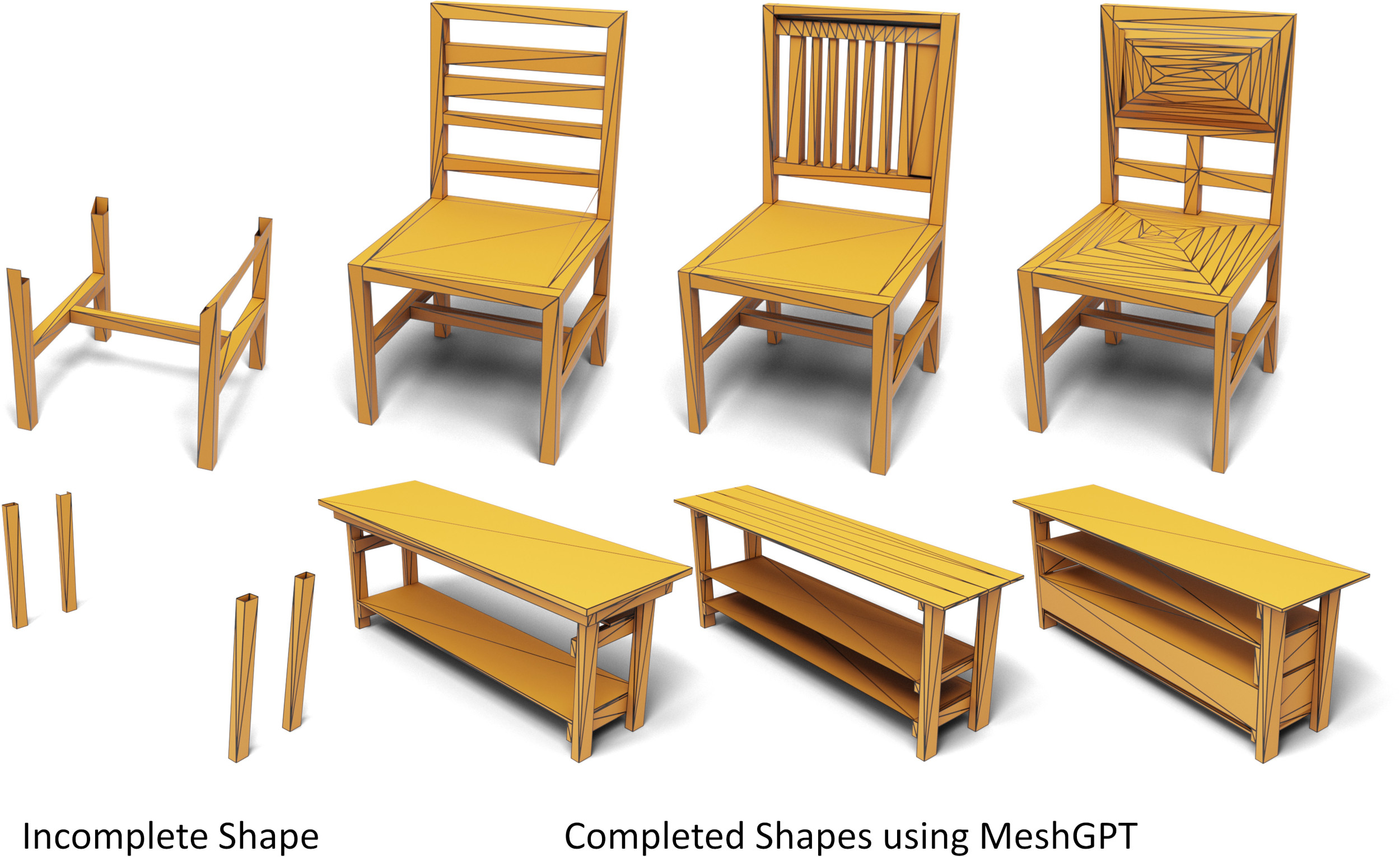}
   \caption{Given a partial mesh, our method can infer multiple possible shape completions.}
   \label{fig:shape_completion}
   \vspace{-5mm}
\end{figure}

%% file: tables/results_user_study.tex
{
\begin{table}[tp]
  \begin{center}
    \small
    \resizebox{\linewidth}{!}{
      \begin{tabular}{lrrrr}
        \toprule
        Preference        & AtlasNet~\cite{groueix2018papier} & BSPNet~\cite{chen2020bsp} & Polygen~\cite{nash2020polygen} & GET3D~\cite{gao2022get3d} \\
        \midrule
        Our Shape         & 82.65\%                           & 78.57\%                   & 85.71\%                        & 68.37\%                   \\
        Our Triangulation & 84.69\%                           & 71.43\%                   & 84.69\%                        & 73.47\%                   \\
        \bottomrule
      \end{tabular}}
    \caption{Percentage of users who prefer our method over the baselines in terms of shape quality and the triangulation quality. Our generated meshes are preferred significantly more often.}
    \vspace{-5mm}
    \label{tab:user_study}
  \end{center}

\end{table}
}

%% file: tables/results_ablation.tex
{
\begin{table}[tp]
  \begin{center}
    \small
    \resizebox{\linewidth}{!}{
      \begin{tabular}{lrrrrr}
        \toprule
        Method                      & COV$\uparrow$  & MMD$\downarrow$ & 1-NNA          & FID$\downarrow$ & KID$\downarrow$  \\
        \midrule
        w/o Learned Tokens          & 27.50          & 4.51            & 93.15          & 40.20           & 0.024            \\
        w/o Encoder Features        & 39.24          & 3.43            & 84.48          & 30.35           & 0.017            \\
        w/o Pretraining             & 36.97          & 3.73            & 84.69          & 27.54           & 0.014            \\
        w/o Sequence Compression    & 30.98          & 4.15            & 88.98          & 38.76           & 0.023            \\
        w/o per Vertex Quantization & 23.57          & 5.49            & 98.35          & 74.94           & 0.050            \\
        \cmidrule{1-6}
        \OURS{}                     & \textbf{43.28} & \textbf{3.29}   & \textbf{75.51} & \textbf{18.46}  & \textbf{0.010} \\
        \bottomrule
      \end{tabular}}
    \caption{Ablations of our design choices on the Chair category of the ShapeNet~\cite{shapenet2015} dataset. As highlighted by the drop in performance by removing any of them, each of these contribute to the final method.}
    \label{tab:ablations}
    \vspace{-7mm}
  \end{center}

\end{table}
}

%% file: figures/ablation.tex
\begin{figure}[htp]
  \centering
   \includegraphics[width=\linewidth]{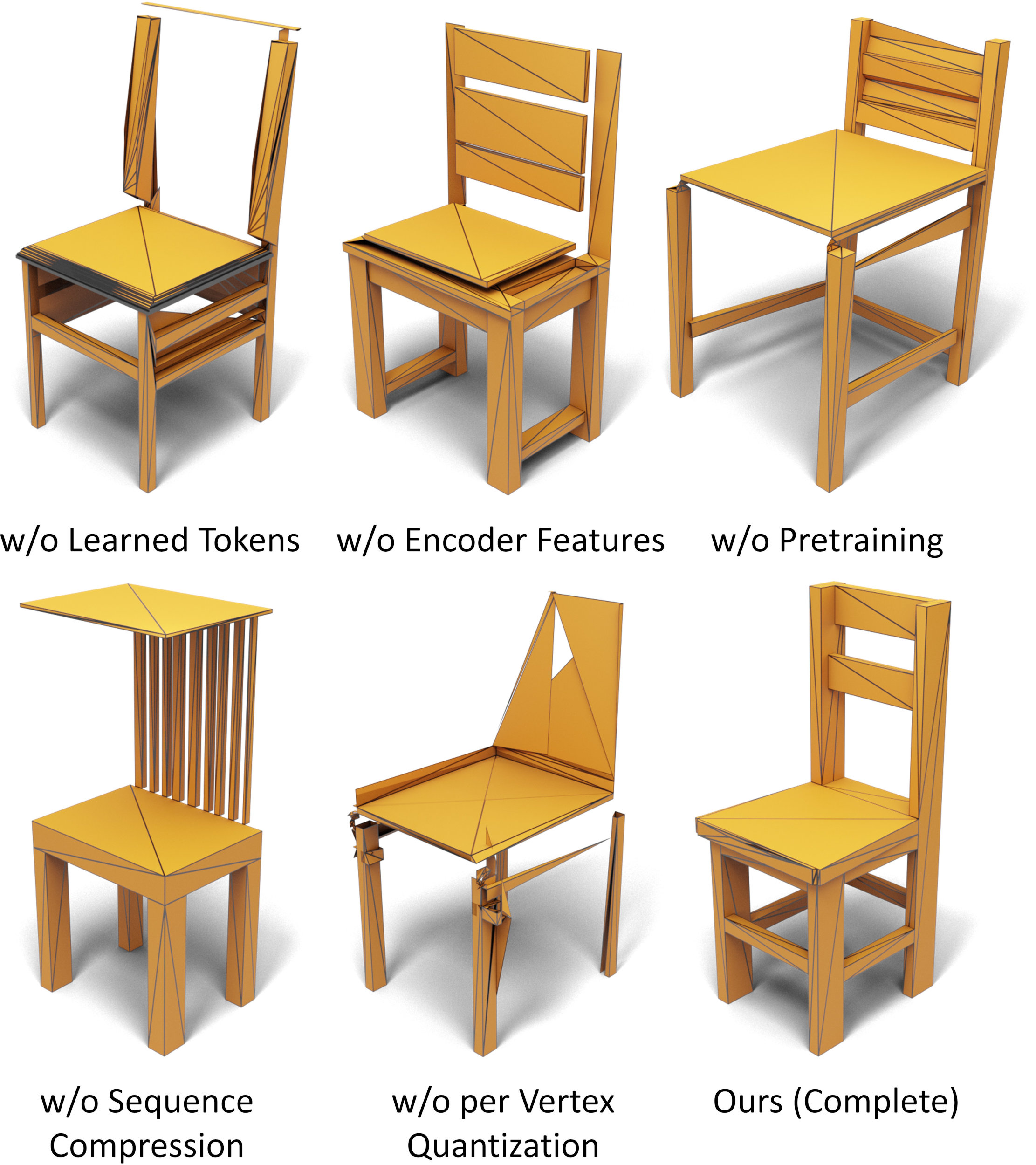}
   \caption{Ablation over our method's components. Naive tokenization (w/o Learned Tokens) and  naive per face quantization (w/o per-vertex quantization) markedly diminishes shape quality. Longer sequences without sequence compression (w/o Sequence Compression) lead to the model forgetting the context and repeating shape elements in the output.}
   \label{fig:results_ablation}
   \vspace{-3mm}
\end{figure}

%% file: sec/5_conclusion.tex
\section{Conclusion}
\label{sec:conclusion}
We have introduced MeshGPT, a novel shape generation approach that outputs meshes directly as triangles. We learn a vocabulary of geometric embeddings over a distribution of meshes, over which a transformer is trained to predict meshes autoregressively as a sequence of triangles.
In contrast to existing mesh generation approaches, our method generates clean, coherent meshes which are compact and follow the triangulation patterns in real data more closely. 
We believe that MeshGPT will not only elevate the current landscape of mesh generation but also inspire new research in the area, offering a unique alternative to the more commonly explored representations for 3D content creation.

%% file: sec/6_acknowledge.tex
\section*{Acknowledgements}
This work was funded by AUDI AG. Matthias Nie{\ss}ner was supported by the ERC Starting Grant Scan2CAD (804724). Angela Dai was supported by the Bavarian State Ministry of Science and the Arts coordinated by the Bavarian Research Institute for Digital Transformation (BIDT).
We would like to thank Ziya Erkoç, Quyet-Chien Nguyen, Haoxuan Li and Artem Sevastopolsky for the helpful discussions.

%% file: sec/7_appendix.tex
In this supplementary document, we discuss additional details about our method \OURS.
We provide implementation details of our method, loss functions, and the baselines in Section~\ref{sec:supp_impldeets}.
Additional details about the user study are provided in Section~\ref{sec:supp_userstudy}.
We also provide further qualitative and quantitative results (Section~\ref{sec:supp_additionalres}), including a shape novelty analysis (Section~\ref{sec:supp_shapennovelty}) for shapes from the main paper.
We further encourage the readers to check out the supplemental video for a summary of the method and an overview of results.

\section{Data}
\label{sec:supp_data}

\mypara{Selection.} We use the ShapeNetV2~\cite{shapenet2015} dataset for all our experiments. We first apply planar decimation to each shape using Blender~\cite{blender}, with the angle tolerance parameter $\alpha$ set within $[1, 60]$. The impact of this decimation is assessed by calculating the Hausdorff distance~\cite{blumberg1920hausdorff} between the decimated and original shapes. We then choose, for each original shape, the decimated version with the Hausdorff distance closest to, but below, a pre-set threshold $\delta_{\text{hausdorff}}$. Shapes with more than 800 faces are excluded, resulting in a final count of 28980 shapes across all categories. The Chair, Table, Bench, and Lamp categories are further divided into a 9:1 train-test split. All shapes from rest of the categories are used for pretraining phase, while only the training subset from specific categories is used for pretraining and finetuning. All shapes are normalized to be centered at the origin and scaled to ensure the longest side is of unit length.

\input{figures/additional_results}
\input{figures/shape_novelty_app_0}
\input{figures/shape_novelty_app_1}

\mypara{Augmentation.} During the training of both the encoder-decoder and the transformer, multiple augmentation techniques are applied to all train shapes. Scaling augmentation, ranging from $0.75$ to $1.25$, is independently applied across each axis. Post-scaling, meshes are resized to keep the longest side at unit length. Additionally, jitter-shift augmentation in the range of $[-0.1, 0.1]$ is used, adjusted to maintain the mesh within the unit bounding box around the origin. We also implement varying levels of planar decimation for training shapes, provided the distortion remains below $\delta_{\text{hausdorff}}$. 

\section{Method Details}
\label{sec:supp_impldeets}

\subsection{Architecture} 
\input{figures/architecture}
The architecture of our encoder-decoder network is elaborated in Fig.~\ref{fig:app_architecture}. The encoder comprises a series of SAGEConv~\cite{hamilton2017inductive} graph convolution layers, processing the mesh in the form of a face graph. For each graph node, input features include the positionally encoded 9 coordinates of the face triangle, its area, the angles between its edges, and the normal of the face. The decoder is essentially a 1D ResNet-34~\cite{he2016deep} network, applied to the face features interpreted as a 1D sequence. It outputs logits corresponding to the 9 discrete coordinates of each face triangle, which are discretized within a $128^3$ space. The codebook $\mathcal{C}$ has a size of $16384$. The architecture of the transformer is simply a GPT-2 medium architecture, i.e. $24$ multi-headed self attention layers, $16$ heads, $768$ as feature width, with context length of $4608$.
\subsection{Residual Vector Quantization}
\mypara{Fundamentals.} For quantization, we employ residual vector quantization (RQ)~\cite{martinez2014stacked, lee2022autoregressive}. RQ discretizes a vector $\mathbf{z}$ with a stack of $D$ ordered codes. Starting with the $0^\text{th}$ residual $\mathbf{r}^0 = \mathbf{z}$, RQ recursively computes $t^d$ as the code of the residual $\mathbf{r}^{d-1}$, and the next residual $\mathbf{r}^{d}$ as
\begin{align}
    t^d &= \mathcal{Q}\,(\mathbf{r}^{d-1}; \mathcal{C})\\
    \mathbf{r}^d &= \mathbf{r}^{d-1} - \mathbf{e}(t^{d})
\end{align} 
where $\mathcal{Q}(\mathbf{z}; \mathcal{C})$ denotes vector quantization of  $\mathbf{z}$ with codebook $\mathcal{C}$, and $\mathbf{e}(t^{d})$ is the embedding in the codebook $\mathcal{C}$. Further, we define
\begin{equation}
\hat{\mathbf{z}}^{(d)} = \sum_{1}^{d}{ \,\mathbf{e}(t^{d})}        
\end{equation}
as the partial sum of up to $d$ code embeddings, and $\hat{\mathbf{z}}=\hat{\mathbf{z}}^{D}$ is the quantized vector of $\mathbf{z}$.
The recursive quantization of RQ thus approximates the vector $\mathbf{z}$ in a coarse-to-fine manner~\cite{lee2022autoregressive}. The commitment loss can now be defined between vector $\mathbf{z}$ and its quantization $\hat{\mathbf{z}}$ as 
\begin{equation}
\mathcal{L}_\text{commit}(\mathbf{z}, \hat{\mathbf{z}}) = \sum_{d=1}^{D} \| \mathbf{z} - \text{sg}[\hat{\mathbf{z}}^{(d)}]\|_{2}^2
\end{equation}
where sg denotes the stop gradient operation.

\mypara{Per Vertex Residual Vector Quantization.} Instead of directly applying RQ, for a face feature $\mathbf{z_i}$ extracted by the graph encoder, we first split this $576$ dimension face feature $\mathbf{z_i}$ into 3 features, $(\mathbf{z}_{i}^1, \mathbf{z}_{i}^2, \mathbf{z}_{i}^3)$, each of $192$ dimensions representing the features of the face triangle's 3 vertices. The features the fall on vertices that are shared across faces are averaged. On these per vertex index feature $\mathbf{z}_{i}^j$, RQ quantizes them into a stack of $\frac{D}{3}$ features,
\begin{equation}
    \text{RQ}(\mathbf{z_i}; \mathcal{C}, D) = (\text{RQ}(\mathbf{z}^{1}_i; \mathcal{C}, \frac{D}{3}), \ldots, \text{RQ}(\mathbf{z}^{3}_i; \mathcal{C}, \frac{D}{3}))
\end{equation}
for codebook $\mathcal{C}$, with 
\begin{equation}
    \text{RQ}(\mathbf{z^{j}_i}; \mathcal{C}, \frac{D}{3}) = (t^{2j - \frac{D}{3} + 1}_i, t^{2j - \frac{D}{3} + 2}_i, \ldots, t^{2j}_i),
\end{equation}
where $t^{d}_i$ is the index to the embedding $\mathbf{e}(t^{d}_i)$ in the codebook $\mathcal{C}$.
Taken together for each vertex, these form a stack of $D$ features, 
\begin{equation}
    \text{RQ}(\mathbf{z_i}; \mathcal{C}, D) = (t_{i}^{1}, t_{i}^{2}, \ldots, t_{i}^{D}) = t_i.
\end{equation}
Thus, the residual quantization for the features extracted for all the $N$ faces of the mesh $\mathbf{Z} = (z_1, z_2, \ldots, z_N)$ is given as
\begin{align}
    \text{RQ}(\mathbf{Z}; \mathcal{C}, D) = \text{RQ}(\mathbf{z_1}\ldots\mathbf{z_N}; \mathcal{C}, D)\\
    \text{RQ}(\mathbf{z_1}\ldots\mathbf{z_N}; \mathcal{C}, D) = (t_0, t_1, \ldots, t_N).
\end{align}
Fig.~\ref{fig:app_sequence_intuition} gives an intuition on why `per vertex' tokenization is better than `per face' tokenization, with ablations in the main paper confirming it.
\subsection{Loss Functions}
\mypara{Vocabulary Learning.} Let $\mathcal{P}_{nijk}$ be the predicted probability distribution over the discrete coordinates, where $n$ is the face index, $i$ is the vertex index inside the face, $j$ is the coordinate's axis index ($x$, $y$ or $z$), and $k$ goes over the discretized positions $\in \{1,2,3,\ldots, 128\}$. If $V_{nij}$ is the target discretized position, then the reconstruction loss for the encoder-decoder network is given as 
\begin{align}
    \mathcal{L}_\text{recon} = \sum_{n=1}^{N}\sum_{i=1}^{3}\sum_{j=1}^{3}\sum_{k=1}^{128} \text{w}_{nijk}\,\operatorname{log}\mathcal{P}_{nijk}
\end{align} with
\begin{align}
    \text{w}_{nijk} = \text{smooth}\,(\text{one-hot}_{128}\,(V_{nij}))
\end{align} 
is a smoothening kernel applied across the one-hot probability distribution over the targets, encouraging physically close coordinates to be penalized less. The loss over the encoder-decoder network is the sum of $\mathcal{L}_\text{recon}$ and $\mathcal{L}_\text{commit}$ previously described.

\mypara{Transformer.} Given a target sequence $\mathbf{T}=(t_0, t_1, \ldots, t_N)$ with $t_i = (t_{i}^{1}, t_{i}^{2}, \ldots, t_{i}^{D})$, and $s_{i}^{j}$ is the corresponding predicted sequence element, then the transformer is trained with the loss
\begin{align}
    \mathcal{L}_\text{recon} = \sum_{i=1}^{N}\sum_{j=1}^{D}\sum_{k=1}^{|\mathcal{C}|}\operatorname{log} p(s_{i}^{k} = t_{i}^{j}).
\end{align}

\subsection{Baselines} We utilize the official implementations for BSPNet~\cite{chen2020bsp}, AtlasNet~\cite{groueix2018papier}, and GET3D~\cite{gao2022get3d}. For Polygen~\cite{nash2020polygen}, we re-implement it following the details in their paper. To align its architecture with our method, we employ the same GPT2-medium architecture for the vertex model in Polygen. Additionally, mirroring our approach, Polygen undergoes pretraining on all categories and is finetuned for each evaluated category, applying the same train-time augmentations as used in our method.

\section{User Study Details}
\label{sec:supp_userstudy}
\input{figures/user_study}
We develop a Django-based web application for the user study. In Fig.~\ref{fig:app_userstudy}, we show the interface for the questionnaire. We randomly select 16 pairs of meshes from each baseline and our method across the Chair and Table categories, half of which are used for a question on preference based on shape quality, and the other half for preference based on triangulation quality. After the samples are prepared, we ask the users to pick the sample which they prefer more based on the question. To avoid biases in this user study, we shuffle the pairs so that there is no positional hint to our method. We also show a collection of ground-truth meshes to the user for them to get an idea of the real distribution. In the end, we gather 784 responses from 49 participants to calculate the preferences.

\section{Shape Novelty Analysis}
\label{sec:supp_shapennovelty}
Fig.~\ref{fig:app_shape_novelty_0} and \ref{fig:app_shape_novelty_1} displays the top-3 most similar shapes from the train set corresponding to all samples used in the main paper that were generated by our model. These nearest neighbor shapes are identified based on Chamfer Distance (CD). To ensure a fair distance computation, accounting for potential discrepancies due to scale or shift in the augmented generations, we normalize all generated and train shapes to be centered within $[0,1]^3$ and scaled to the extremes of this cube.

\section{Additional Results}
\label{sec:supp_additionalres}
\mypara{Metrics.} Following recent works for unconditional shape generation~\cite{yu2022point, zeng2022lion, erkocc2023hyperdiffusion} for calculating the shape metrics we define

\begin{align*}
\text{MMD}(S_g, S_r) &= \frac{1}{\vert S_r \vert} \sum_{Y \in S_r} \min_{X \in S_g} D(X, Y), \\
\text{COV}(S_g, S_r) &= \frac{\vert \{ \argmin_{Y \in S_r} D(X, Y) \vert X \in S_g \} \vert}{\vert S_r \vert}, \\
\text{1-NNA}(S_g, S_r) &= \frac{\sum_{X \in S_g} \mathbbm{1}_X + \sum_{Y \in S_r} \mathbbm{1}_Y }{\vert S_g \vert + \vert S_r \vert}, \\
\mathbbm{1}_X &= \mathbbm{1}[N_X \in S_g], \\
\mathbbm{1}_Y &= \mathbbm{1}[N_Y \in S_r],
\end{align*}
where in the 1-NNA metric $N_X$ is a point cloud that is closest to $X$ in both generated and reference dataset, i.e., 
$$N_X = \argmin_{K \in S_r \cup S_g} D(X, K)$$
We use a Chamfer Distance (CD) distance measure $D(X,Y)$ for computing these metrics in 3D. To evaluate these point-based measures, we sample $2048$ points randomly from all baseline outputs; and use $6000$, $1200$, $1000$, $8000$ generated shapes from chair, bench, lamp and table categories.

\mypara{Qualitative Results.} Fig.~\ref{fig:additional_results} shows more unconditional generations from our model across different ShapeNet categories.

\input{tables/results_encdec}
\mypara{Encoder-Decoder Ablations.} In Tab.~\ref{tab:app_ablations}, we show a set of ablations on the design choice for our encoder-decoder network used for learning the triangle embeddings. We measure the performance in terms of triangle accuracy, which measures average accuracy with which all 9 coordinates of faces are correctly predicted, and the cross-entropy loss on the test set. 

\input{figures/sequence_intuition}

We evaluate the effect of various choices -- how much does the positional encoding at input help, effect of using continuous predictions instead of discrete as outputs, using vector quantization (1 token per face) instead of residual quantization ($D$ tokens per face), encoder architecture as a point encoder, or different graph convolution operators, and decoder architecture as either ResNet19 or PointNet decoder. Note that even though for encoder-decoder reconstruction, `w/o per Vertex Quantization' performs best, this variant works significantly worse than with per Vertex Quantization, as shown in the main paper. Fig.~\ref{fig:app_sequence_intuition} describes an intuition of why the embeddings from this variant are more transformer friendly.

%% file: figures/additional_results.tex
\begin{figure*}[htp]
  \centering
   \includegraphics[width=0.85\linewidth]{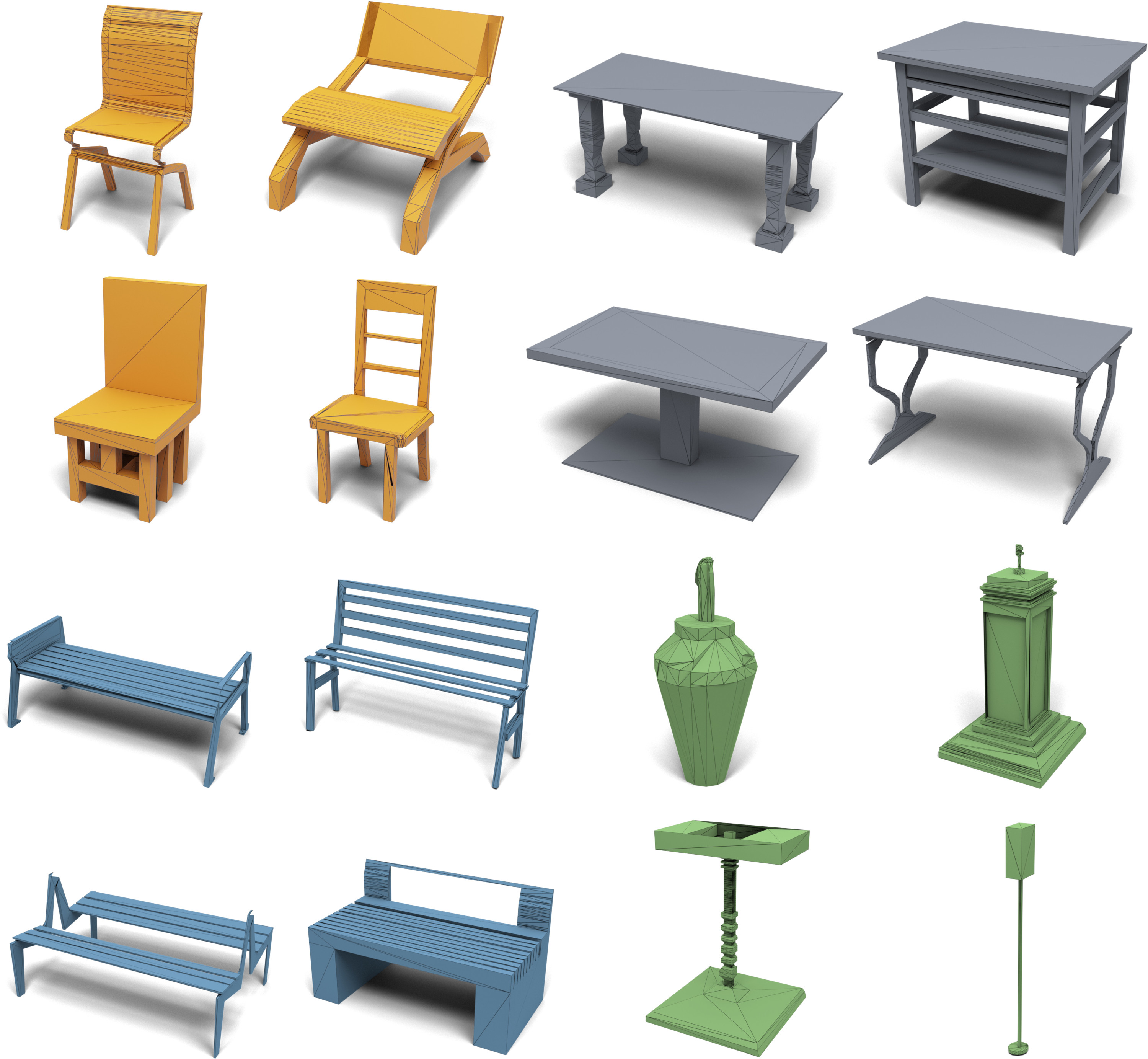}
   \caption{Additional novel shapes on Chairs, Tables, Benches and Lamps generated by our method.}
   \label{fig:additional_results}
\end{figure*}

%% file: figures/shape_novelty_app_0.tex
\begin{figure*}[htp]
  \centering
   \includegraphics[width=\linewidth]{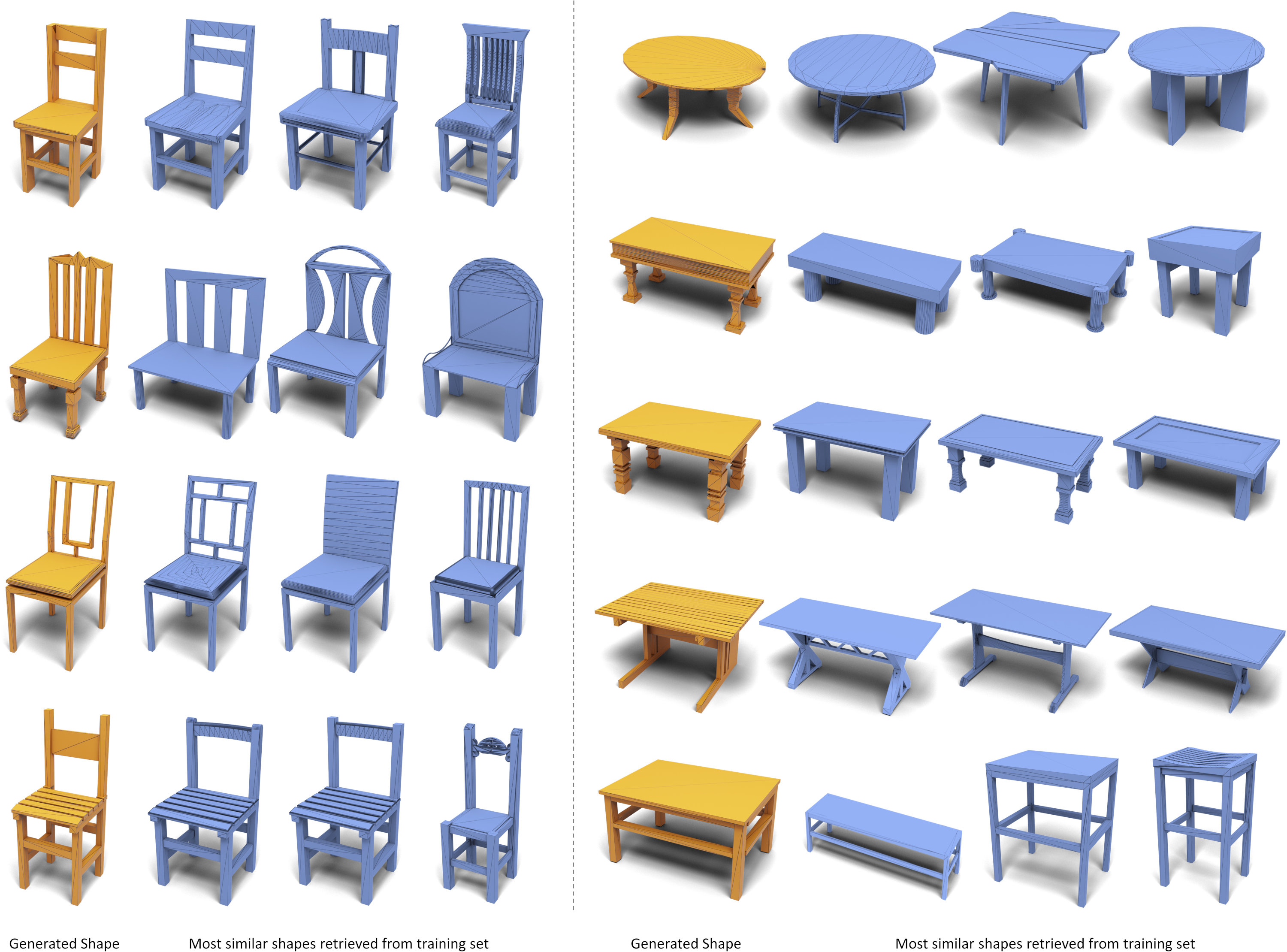}
   \caption{Shape novelty analysis on ShapeNet~\cite{shapenet2015} chair and table category for shapes generated by our method shown in main paper. We show the 3 nearest neighbors in terms of Chamfer Distance (CD) for a generated shape. Shapes are scaled to a unit length along all axes to account for augmented generations before computing CD.}
   \label{fig:app_shape_novelty_0}
\end{figure*}

%% file: figures/shape_novelty_app_1.tex
\begin{figure*}[htp]
  \centering
   \includegraphics[width=\linewidth]{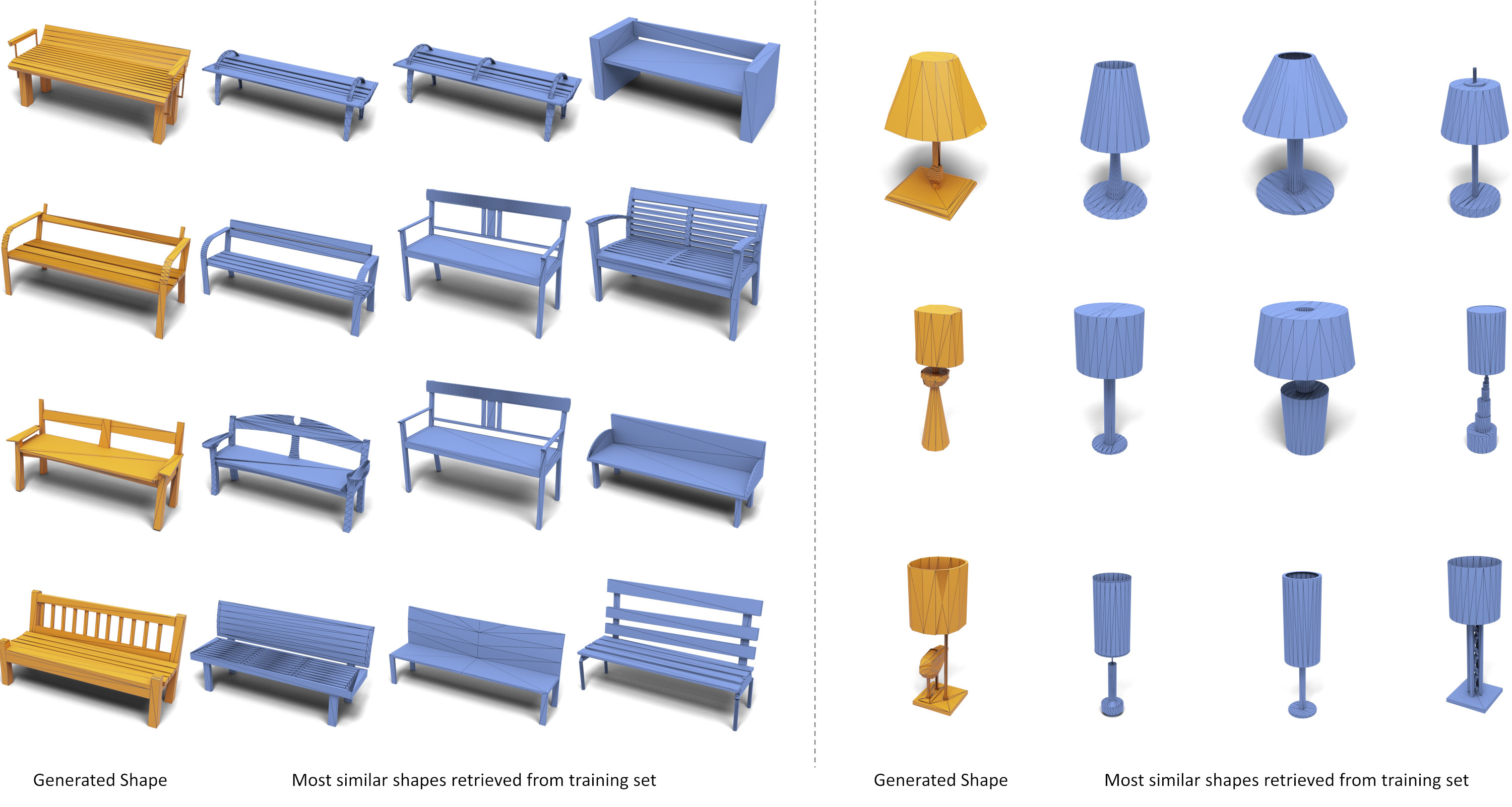}
   \caption{Shape novelty analysis on ShapeNet~\cite{shapenet2015} bench and lamp category for shapes generated by our method shown in main paper. We show the 3 nearest neighbors in terms of Chamfer Distance (CD) for a generated shape. Shapes are scaled to a unit length along all axes to account for augmented generations before computing CD.}
   \label{fig:app_shape_novelty_1}
\end{figure*}

%% file: figures/architecture.tex
\begin{figure}[htb]
  \centering
   \includegraphics[width=\linewidth]{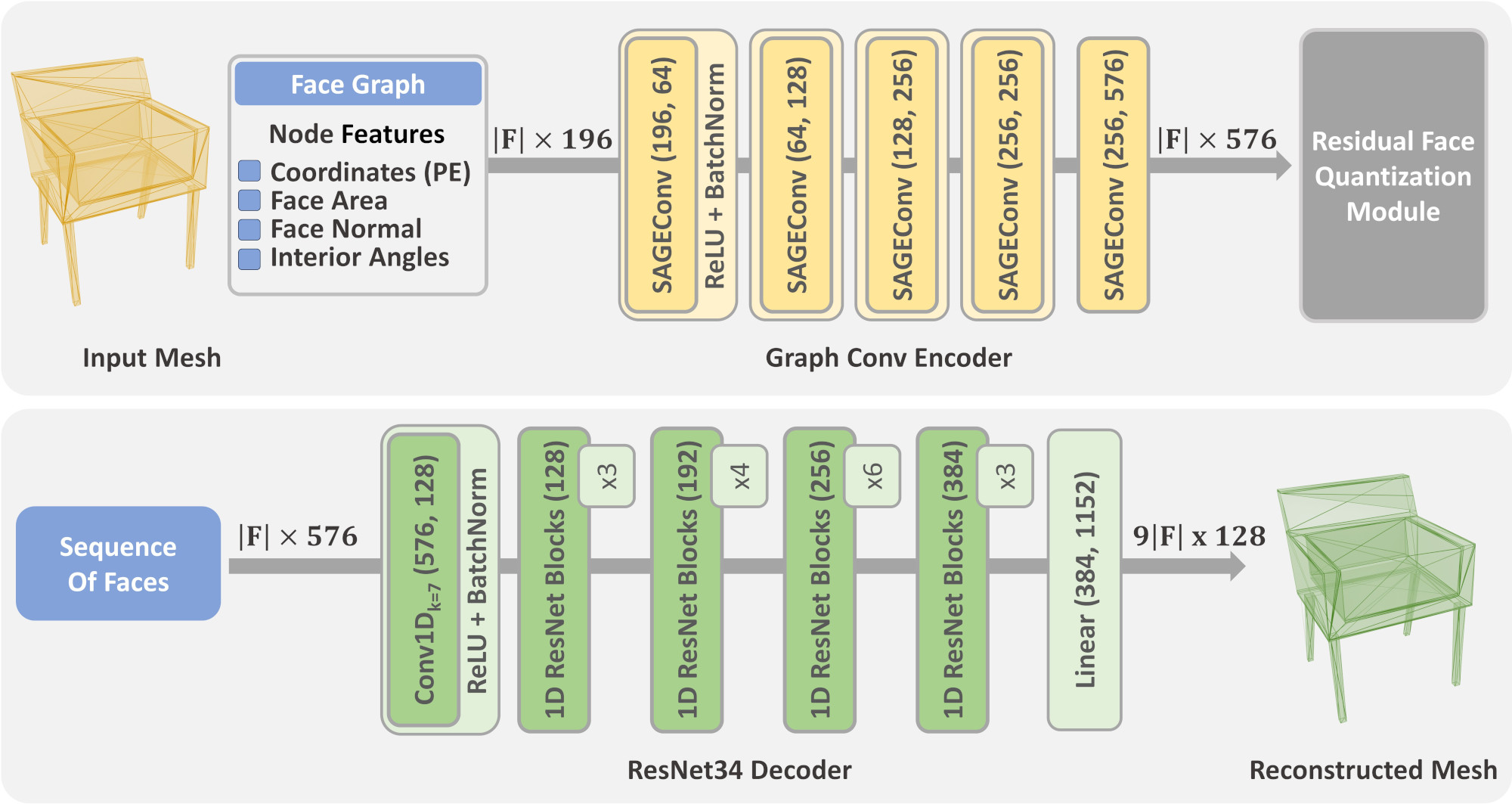}
   \caption{
   Our encoder-decoder network features an encoder with SAGEConv~\cite{hamilton2017inductive} layers processing mesh faces as a graph. Each node inputs positionally encoded face triangle coordinates, area, edge angles, and normal. The decoder, a 1D ResNet-34~\cite{he2016deep}, interprets face features as a sequence, outputting logits for the discretized face triangle coordinates in a $128^3$ space.
   }
   \vspace{-3mm}
   \label{fig:app_architecture}
\end{figure}

%% file: figures/user_study.tex
\begin{figure}[htp]
  \centering
   \includegraphics[width=\linewidth]{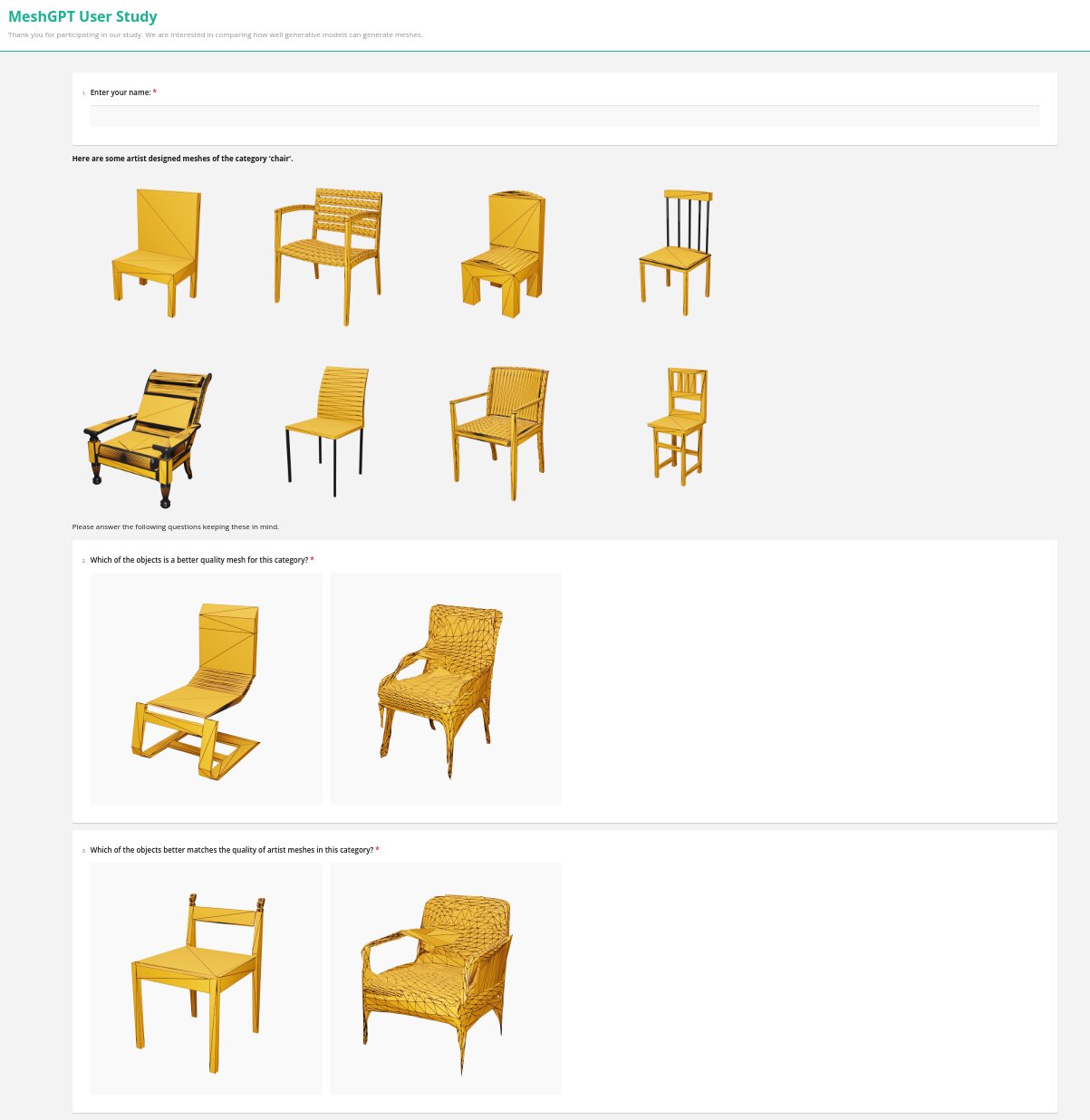}
   \caption{
    User study interface. We show users a set of random ground-truth shapes for a category and then ask users for shape quality and triangulation preference among meshed generated by two methods.  
   }
   \label{fig:app_userstudy}
\end{figure}

%% file: tables/results_encdec.tex
{
\begin{table}[tp]
  \begin{center}
    \small
    \resizebox{\linewidth}{!}{
      \begin{tabular}{lrr}
        \toprule
        Variant                             & Triangle Accuracy (\%) $\uparrow$ & Cross-Entropy $\downarrow$ \\
        \midrule
        w/o Positional Encoding             & 79.33                       & 0.2484                    \\
        w/o Output Discretization           & 22.03                       & 0.5705                    \\
        w/o Residual Quantization           & 1.29                        & 4.6679                      \\
        w/o per Vertex Quantization         & \textbf{98.64}                       & \textbf{0.1413} \\
        w/ PointNet Encoder                 & 88.73                       & 0.1896                    \\
        w/ GAT~\cite{velivckovic2017graph} Encoder           & 86.14                       & 0.2015                    \\
        w/ EdgeConv~\cite{wang2019dynamic} Encoder & 91.23                       & 0.1702                    \\
        w/ ResNet19 Decoder                 & 96.29                       & 0.1492                    \\
        w/ PointNet Decoder                 & 95.47                       & 0.1528                    \\
        \cmidrule{1-3}
        \OURS{}                             & 98.49                      & 0.1473                    \\
        \bottomrule
      \end{tabular}}
    \caption{Ablations of our design choices for the encoder-decoder network on the Chair category of the ShapeNet~\cite{shapenet2015} dataset.}
    \label{tab:app_ablations}
    \vspace{-7mm}
  \end{center}

\end{table}
}

%% file: figures/sequence_intuition.tex
\begin{figure}[htp]
  \centering
   \includegraphics[width=\linewidth]{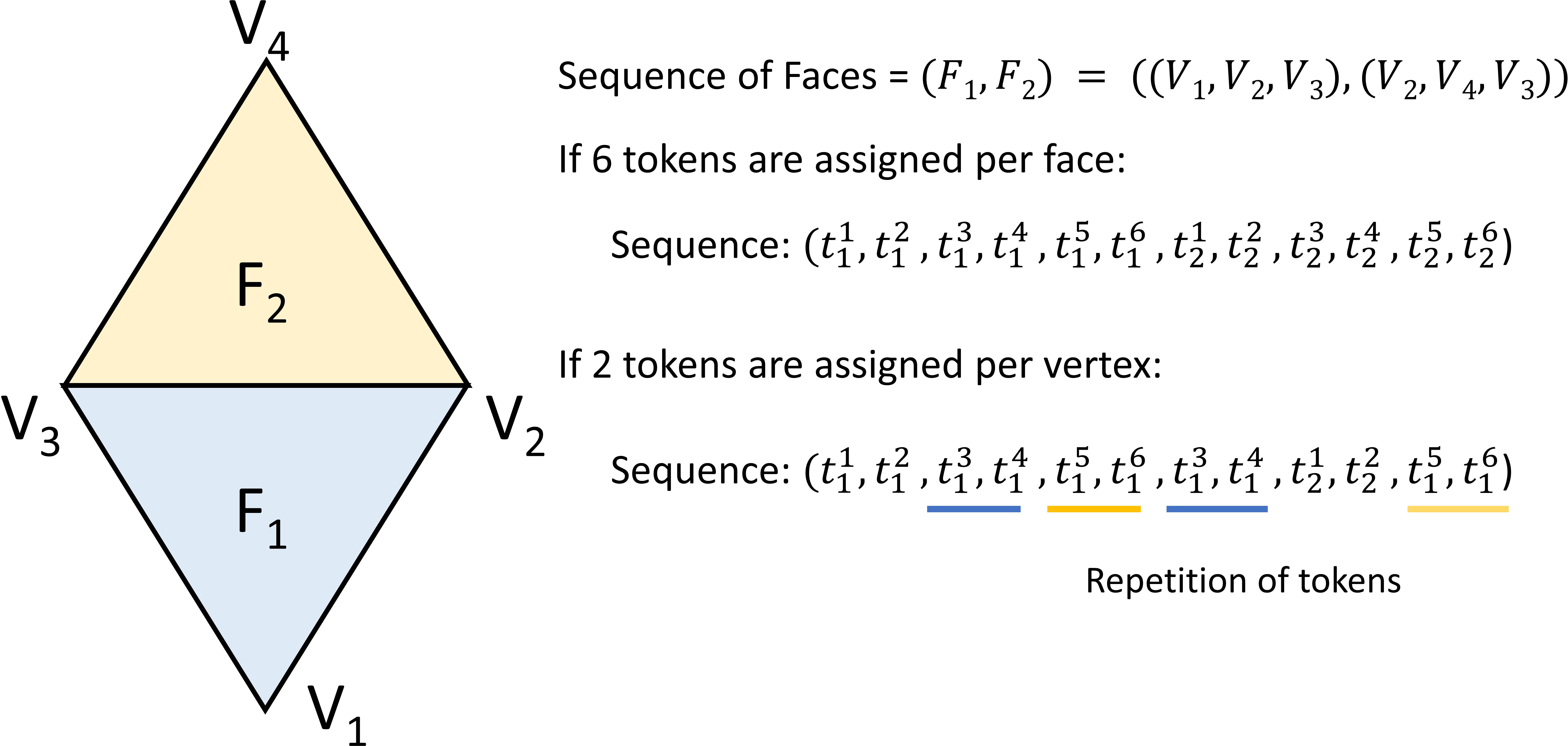}
   \caption{The effectiveness of per-vertex quantization over per-face quantization can be understood through an example where two faces share an edge as shown above. With per-face tokenization assigning 6 tokens per face, the sequence yields 12 unique tokens. In contrast, per-vertex tokenization leads to repeated tokens in the sequence due to shared vertices between faces. This repetition makes the sequence easier for the transformer to learn compared to a wholly unique sequence per face, especially when both sequences are of equal length.}
   \label{fig:app_sequence_intuition}
\end{figure}